\documentclass[a4paper, 11pt]{article}

\usepackage[numbers]{natbib}
\usepackage[a4paper, margin=2.5cm]{geometry}

\usepackage{amsmath,amsfonts}
\usepackage{algorithmic}
\usepackage{algorithm}
\usepackage{array}
\usepackage{textcomp}
\usepackage{stfloats}
\usepackage{url}
\usepackage{verbatim}
\usepackage{graphicx}
\usepackage{boxedminipage}
\usepackage{amssymb}
\usepackage{amsthm}
\usepackage{t1enc}
\usepackage{svg}
\usepackage{caption}
\usepackage{subcaption}
\usepackage{tikz}
\usepackage{float}
\usepackage{bbm}
\usepackage{authblk}

\newtheorem{theorem}{Theorem}
\newtheorem{example}{Example}%
\newtheorem{definition}{Definition}

\newcommand{\floor}[1]{\left\lfloor #1 \right\rfloor}

\begin{document}
\let\WriteBookmarks\relax
\def\floatpagepagefraction{1}
\def\textpagefraction{.001}

\title{pbn-STAC: Deep Reinforcement Learning-based Framework for Cellular Reprogramming}

\author[1, 2]{Andrzej Mizera}
\author[1, 2]{Jakub Zarzycki}

\affil[1]{University of Warsaw}
\affil[2]{IDEAS NCBR}

\maketitle

\begin{abstract}
Cellular reprogramming can be used for both the prevention and cure of complex diseases. However, the efficiency of discovering reprogramming strategies with classical wet-lab experiments is hindered by lengthy time commitments and high costs. In this study, we leverage deep reinforcement learning to develop a~novel computational framework that facilitates the identification of reprogramming strategies. To this aim, we formulate a~control problem in the context of cellular reprogramming for the Boolean and probabilistic Boolean network models of gene regulatory networks under the asynchronous update mode. Furthermore, to facilitate scalability, we introduce the notion of a~pseudo-attractor and a~procedure for the identification of pseudo-attractor states. Finally, we devise a~computational framework for solving the control problem, which we test on a~number of biological networks. 
\end{abstract}



\section{Introduction}
Complex diseases pose a~great challenge largely because genes and gene products operate within a~complex system --~the {\it gene regulatory network} (GRN). There is an~inherent dynamic behaviour emerging from the structural wiring of a~GRN: gene expression profiles, i.e., states of a~GRN, evolve in time to finally reach stable states referred to as {\it attractors}. Attractors correspond to cell types or cell fates~\citep{HEBI05}. During normal development of a~multi-cellular organism, not all attractors are manifested. Some of the `abnormal attractors', associated with diseases, become accessible by disturbance of the GRN's dynamics. This is seldom a~consequence of a~disruption in a~single gene, but rather arises as an~aftermath of GRN perturbations~\citep{Barabasi2011NetMed}. This could be cured by guiding cells to desired `healthy' attractors with experimental procedures of {\it cellular reprogramming}.
Unfortunately, finding effective interventions that trigger desired changes using solely wet-lab experiments is difficult, costly, and requires lengthy time commitments. This motivates us to consider \emph{in-silico} approaches.
In the context of computational modelling of GRNs, cellular reprogramming can be stated as a~control problem. Since GRNs are complex systems characterised by emergence, computing cellular reprogramming procedures requires large GRN models to be considered. The modelling frameworks of Boolean networks (BNs) and probabilistic Boolean networks (PBNs) are well-suited for the holistic modelling of GRNs due to the advantage of being simplistic generalisations of the system's dynamics and yet capable of capturing the intrinsic non-linear dynamic properties. 

Computation of proper control strategies for non-linear systems requires both network structure and dynamics~\citep{GR16}. Recently, a~number of state-of-the-art methods have been proposed. An~efficient method based on the `divide and conquer' strategy was proposed in~\citep{PSPM20} to solve the minimal {\em one-step source-target control} problem by using instantaneous, temporary, and permanent gene perturbations. The minimal {\em sequential source-target control} and the {\em target control} problems of BNs were considered in~\citep{SP20a} and~\citep{SP20b}, respectively. All these methods were implemented in the CABEAN~\citep{SP21} tool. Recently, semi-symbolic algorithms were proposed in~\citep{BBPS+23} to stabilise partially specified asynchronous BNs in states exhibiting specific traits. In~\citep{Pauleve23BoNesis}, the control problem for the most permissive BN update mode in the context of fixed points and minimal trap spaces was considered.

The existing structure- and dynamics-based state-of-the-art computational techniques are limited to small and mid-size networks due to the infamous state space explosion problem. They usually require the systems to facilitate some kind of  decomposition. This however is often too restrictive for cellular reprogramming considerations. 

The issue of scalability can be addressed by devising new methods based on deep reinforcement learning (DRL) techniques, which have proved very successful in decision problems characterised by huge state-action spaces.  The application of reinforcement learning for controlling GRNs was pioneered in~\citep{SPA13} with focus on how to control GRNs by avoiding undesirable states in terms of steady state probabilities of PBNs. The main idea was to treat the time series gene expression samples as a~sequence of experience tuples and use a~batch version of Q-Learning to produce an~approximated policy over the experience tuples. Later, the BOAFQI-Sarsa method that does not require time series samples was devised in~\citep{NCB18}. A~batch reinforcement learning method, mSFQI, was proposed in~\citep{NBC20} for control based on probabilities of gene activity profiles. Recently, the study of~\citep{AYGV20} used a~Deep Q-Network with prioritised experience replay, for control of synchronous PBNs to drive the networks from a~particular state towards a~more desirable one. Finally, a~DRL-based approximate solution to the control problem in synchronous PBNs was proposed in~\citep{MCSW22}. The proposed method finds a~control strategy from any network state to a~specified target attractor using a~Double Deep Q-Network.

This study represents an~exploration into the potential of DRL for developing scalable control methods for both BN and PBN models of GRNs in the context of cellular reprogramming. 
Our contributions are as follows. We formulate a~novel control problem, i.e., the source-target attractor control for BN/PBN models of GRNs, tailored for cellular reprogramming in the sense that control interventions are allowed to take place only in states corresponding to phenotypic states of the cell. Moreover, in contrast to the existing studies on DRL-based control methods, where synchronous update scheme was considered, our focus is on the computationally more challenging asynchronous mode since it is commonly considered more appropriate for the modelling of GRNs. Next, as identifying attractors in large networks is ineffective by itself, we introduce the notion of a~pseudo-attractor and develop a~procedure for identifying pseudo-attractor states during DRL agent training. We then devise pbn-STAC, i.e., a~DRL-based framework for solving the source-target attractor control problem, which we test on a~number of case studies. In order for the predictions on the control strategies to be more realistic in terms of wet-lab practices, the control actions are restricted to be applied in (pseudo-)attractor states only.
We consider our contributions as a~relevant step towards development of scalable computational methods based on DRL for the control of BN/PBN models of GRNs in the context of cellular reprogramming.
We present the materials, methods, and experimental results of pbn-STAC in the subsequent sections.

\section{Materials and methods}
\subsection{Boolean networks}
A~PBN is a~discrete dynamical system defined as a~pair $(V, \mathcal{F})$, where $V = \{x_1, x_2, \ldots, x_n\}$ is a~set of binary-valued nodes (also referred to as genes) and $\mathcal{F} = (F_1, F_2,\ldots, F_n)$ is a~list of sets. Each node $x_i \in V$ has associated a~set $F_i \in \mathcal{F}$ of predictor functions, i.e., $F_i = \{f^i_1, f^i_2, \ldots, f^i_{l(i)}\}$, where $l(i)$ is the number of predictor functions of node $x_i$. Each $f^i_j \in F_i$ is a~Boolean function defined with respect to a~subset of $V$ denoted $\textrm{Pa}(f^i_j)$ and referred to as parent nodes of $f^i_j$. For each $x_i \in V$, there is a~probability distribution $\mathbf{c}^i = (c^i_1, c^i_2, \ldots, c^i_{l(i)})$ on $F_i$, where each predictor function $f^i_j \in F_i$ has an~associated selection probability denoted $c^i_j$; it holds that $\sum_{j=1}^{l(i)}c^i_j=1$. A~BN is a~PBN in which each node admits only a~single predictor function.

We define a~\emph{state} of a~BN/PBN as an~$n$-dimensional vector $\mathbf{s} \in \{0,1\}^n$, where the $i$-th element represents the state of gene $x_i \in V$. A~BN/PBN evolves by starting in an~initial state and updating it at discrete time steps in accordance with the selected predictor functions. Herein, we focus on the asynchronous updating scheme, where at each time step the state value for a~single gene $x_i$ is updated in accordance with a~randomly selected predictor function from $F_i$ in accordance with $\mathbf{c}^i$. 

The resulting network dynamics can be represented in the form of a~\emph{state transition graph} (STG). An~STG of a~BN/PBN of $n$ genes under the asynchronous update mode is a~graph $STG(S,\rightarrow)$, where $S = \{0,1\}^n$ is the set of all possible states and $\rightarrow$ is the set of directed edges such that a~directed edge from $s$ to $s'$, denoted $s \rightarrow s'$, is in $\rightarrow$ if and only if $s'$ can be obtained from $s$ by a~single asynchronous update. An~\emph{attractor} of a~BN/PBN is a~bottom strongly connected component in the STG of the network. A~\emph{fixed-point attractor} and a~\emph{multi-state attractor} are bottom strongly connected components consisting of a~single state or more than one state, respectively. We refer to Appendix~\ref{app:BN_PBN} for a~PBN example illustrating the notions mentioned in this section.

\subsection{Reinforcement Learning}

The main task of reinforcement learning (RL) is to solve sequential decision problems by maximising a~cumulative reward.
A~\emph{policy} is a~strategy that determines which action to take and an~\emph{optimal policy} is one determined by selecting the actions that maximise the future cumulative reward. The optimal policy can be obtained by solving the Bellman equation for the $Q$ function which is defined as the total discounted reward received after taking action $a$ in state~$s$.

In the case of large state-action spaces, the $Q$ function values often cannot be determined, therefore they are approximated by training a~ DRL agent. It was shown that as the agent explores the environment, this approximation converges to the true $Q$ values. We refer to Appendix~\ref{app:rl} for a~more detailed introduction to RL and DRL; see~\citep{rl} for a~comprehensive treatment of these subjects.

\subsection{Pseudo-attractors}
Obtaining the attractor landscape of a~BN network, i.e., the family of all its attractors, is an~NP-hard problem by itself~\citep{Akutsu2003}. Although many attractor identification algorithms have been proposed in the literature, they can only handle networks with up to a~few dozens of nodes~(\citep{Mori2022}). Hence, one cannot expect to be in possession of the information on all the attractors prior to solving the control problem for large network models. However, as we argue in Sec.~\ref{sec:algorithm}, knowing the attractor states is crucial in the context of our control problem for cellular reprogramming, where for the sake of biological realism, interventions are allowed to take place only in states corresponding to phenotypic states of the cell. Therefore, to address this issue, we devise our computational framework for the control of BN/PBN models based on DRL to be able to identify the relevant attractor states during training, i.e., the exploration of the DRL environment, without the need of identifying the attractors in advance. 


In the case of classical PBNs, attractors correspond to the irreducible sets of states in the underlying Markov chain~\citep{PBN02}. For large-size PBNs with different predictors for numerous individual genes, the multi-state attractors may be large. Nevertheless, usually states of an~irreducible set are not revisited with the same probability. From the point of view of the control problem in the context of cellular reprogramming, only the frequently revisited states of an~attractor are the relevant ones since they correspond to phenotypical cellular states that are observable in the lab. This makes these states `recognisable' for the application of reprogramming interventions in practice in accordance with the control strategy obtained with our framework. We refer to the subset of frequently revisited states of an~attractor as a~\emph{pseudo-attractor} associated with the attractor.

\begin{definition}[Pseudo-attractor]
\label{def:pseudo-attractor}
Let $A$ be an~attractor of a~PBN, i.e., an~irreducible set of states of the Markov chain underlying the PBN. Let $n := |A|$ be the size of the attractor $A$ and let $\mathbb{P}_A$ be the unique stationary probability distribution on $A$. The \emph{pseudo-attractor} associated with $A$ is the maximal subset $PA \subseteq A$ such that for all $s \in PA$ it holds that $\mathbb{P}_A(s) \geq \frac{1}{n}$. The states of a~pseudo attractor are referred to as \emph{pseudo-attractor states}.
\end{definition}

The correctness of the definition is guaranteed by the fact that the state space of the underlying Markov chain of a~PBN is finite and that the Markov chain restricted to the attractor states is irreducible. It is a~well known fact that all states of a~finite and irreducible Markov chain are positive recurrent. In consequence, the attractor restricted Markov chain has a~unique stationary distribution. Furthermore, for any PBN attractor there exists a~non-empty pseudo-attractor as stated by the following two theorems.

\begin{theorem}
\label{th:existence}
Let $A$ be an~attractor of a~PBN. Then there exists a~pseudo-attractor $PA \subseteq A$ such that $|PA| \geq 1$.     
\end{theorem}

\ifx\noproof\undefined
\begin{proof}
Let $n$ be the size of the attractor $A$, i.e., $n:=|A|$. Since the underlying Markov chain of the PBN restricted to $A$ is irreducible and positive recurrent, it has a~unique stationary distribution, which we denote $\mathbb{P}_A$. We proceed to show that there exists at least one state $s' \in A$ such that $\mathbb{P}_A(s') \geq \frac{1}{n}$. For this, let us assume that no such state exists. Then, we have that
$\sum_{s \in A}\mathbb{P}_A(s) < \sum_{s\in A}\frac{1}{n} = n\cdot\frac{1}{n} = 1$, i.e., the leftmost sum is strictly less than 1. Thus, $\mathbb{P}_A$ is not a~probability distribution on $A$, which leads to a~contradiction, and $|PA| \geq 1$.
\end{proof}
\fi

\begin{theorem}
\label{th:uniform}
Let $A$ be an~attractor of a~PBN such that the unique stationary distribution of the underlying Markov chain of the PBN restricted to $A$ is uniform. Then, for the pseudo-attractor $PA$ associated with $A$ it holds that $PA=A$, i.e., the pseudo-attractor and the attractor are equal. 
\end{theorem}

\ifx\noproof\undefined
\begin{proof}
Let $n$ be the size of $A$. By the assumption of uniformity of the stationary distribution $\mathbb{P}_A$ it holds that $\mathbb{P}_A(s) = \frac{1}{n}$ for each $s \in A$. Since $PA$ is the maximal subset of $A$ such that $\mathbb{P}_A(s) \geq \frac{1}{n}$ for each $s \in A$, it follows that $PA=A$.
\end{proof}
\fi

Notice that Def.~\ref{def:pseudo-attractor} straightforwardly extends to asynchronous BNs and that Th.~\ref{th:existence} and Th.~\ref{th:uniform} remain valid in this case. Indeed, the asynchronous dynamics on a~multi-state BN attractor is a~finite and irreducible Markov chain. Therefore, in the continuation, we use the notion of the pseudo-attractor both in the context of PBNs and BNs.

\subsection{Control problem formulation}
\label{sec:control_problem}

Herein we formulate a~novel control problem for BN/PBN models of GRNs tailored for cellular reprogramming understood as the process of changing the identity or functional state of a~cell by altering its gene expression profiles. We start by providing the~definition of an~\emph{attractor-based control strategy}, also referred to as \emph{control strategy} for short.

\begin{definition}[Attractor-based Control Strategy]
\label{def:control_strategy}
Given a~BN/ PBN and a~pair of its source-target (pseudo-)attractors, an~attractor-based control strategy is a~sequence of interventions which drives the network dynamics from the source to the target (pseudo-)attractor. Interventions are understood as simultaneous flips (perturbations) of values for a~subset of genes in a~particular network state and their application is limited to (pseudo-)attractor states.
Furthermore, the \emph{length of a~control strategy} is defined as the number of interventions in the control sequence. We refer to an~attractor-based control strategy of the shortest length as the \emph{minimal attractor-based control strategy}.
\end{definition}

We now proceed to define our new control problem.

\begin{definition}[Source-Target Attractor Control]
Given a~BN/ PBN and a~pair of source-target (pseudo-)attractor states, find a~minimal attractor-based control strategy.
\end{definition}

Examples illustrating these notions can be found in Appendix~\ref{app:BN_PBN}. Notice that our definition of the source-target attractor control is a~generalisation of the `attractor-based sequential instantaneous control (ASI)' problem defined for BNs in~\citep{SP20a} in the sense that our formulation extends to the formalism of PBNs and pseudo-attractor states. An~exact `divide-and-conquer'-type algorithm for the ASI problem is implemented in CABEAN~\citep{SP21}.

\section{Algorithm}
\label{sec:algorithm}
We propose pbn-STAC --~a~DRL-based computational framework for the source-target attractor control. Since our control problem is to some extent similar to the one considered in~\citep{MCSW22} and our implementation is based on the implementation therein, we compare our framework assumptions and solutions to theirs during the presentation of pbn-STAC. In contrast to the synchronous PBN update mode in~\citep{MCSW22}, we consider the asynchronous update mode. The approach of~\citep{MCSW22} allows DRL agents to apply control actions in any state of the PBN environment. Since our focus is on the modelling of cellular reprogramming, we believe that this approach may be hard to apply in experimental practice. It would require the ability to discern virtually all cellular states, including the transient ones, which is impossible with currently available experimental techniques. Since attractors correspond to cellular types or, more generally, to cellular phenotypic functional states, which are more easily observable in experimental practice, we allow our DRL agent to intervene only in (pseudo-)attractor states in consistency with the control problem formulation in Sec.~\ref{sec:control_problem}.

In the framework of~\citep{MCSW22}, an~action of the DRL agent can perturb at most one gene at a~time. However, for our formulation of the control problem this is too restrictive. We have encountered examples of source-target attractor pairs where no control strategy consisting of such actions exists. Therefore, we need to relax this restriction. However, we do not want to intervene on too many genes at once as it would be rather pointless --~in the extreme case of allowing all genes to be perturbed at once, one could simply flip all of the unmatched gene values. Furthermore, such an~intervention would also be hard to realise or even be unworkable in real biological scenarios --~it is expensive and sometimes even impossible to intervene on many genes at once in the lab. Hence, we introduce a~parameter which determines an~upper limit for the number of genes that can be simultaneously perturbed. Based on experiments (data not shown), we set its value to three. This setting suffices to obtain successful control strategies for all of our case studies, yet low enough not to trivialise the control problem. Of course, the value can be tuned to meet particular needs.

The DRL agent in~\citep{MCSW22} learns how to drive the network dynamics from any state to the specified target attractor. With the context of cellular reprogramming in mind, we consider in our framework only attractors as control sources and targets with both of them specified. This models the process of transforming a~cell from one type into another. We define the DRL reward function as
$R_a(s, s') = 1000 * \mathbbm{1}_{TA}(s') - |a|$, where $s$ and $s'$ are the current and next attractor states, respectively, $\mathbbm{1}_{TA}$ is an~indicator function of target attractor, and $|a|$ is the number of genes perturbed by applying action $a$. The idea behind this reward scheme is to penalise for taking too many actions yet to balance the penalties with a~substantial award after achieving the target state. We use the Double Deep Q Network (DDQN) loss function as in~\citep{MCSW22}. For each DRL episode, we aim to randomly choose a~source-target attractor pair and terminate each episode after 20 unsuccessful actions. This approach however would require all the attractors to be known prior to training. Unfortunately, as already mentioned, obtaining the attractor landscape for large models is difficult. To address this issue, we have introduced the notion of a~pseudo-attractor in Def.~\ref{def:pseudo-attractor} and, instead of attractors, we use the notion of pseudo-attractor states to define the DRL episode of our framework. We now proceed to present a~procedure for detecting pseudo-attractor states which is exploited by pbn-STAC for solving the control problem for large networks.

\subsection{Identification of pseudo-attractor states}
\label{sec:pas-procedure}

Identification of pseudo-attractors is hindered in large models. Nevertheless, pseudo-attractor states can be identified with simulations due to the property of being frequently revisited. We propose the following \emph{Pseudo-Attractor States Identification Procedure} (PASIP) which consists of steps executed in two phases: (I) environment pre-processing and (II) DRL agent training.

\paragraph{\textbf{Pseudo-Attractor States Identification Procedure}}
\begin{enumerate}
  \item[I] During environment pre-processing, a~pool of $k$ randomly selected initial states is considered, from which PBN simulations are started. Each PBN simulation is run for initial $n_0=200$ time steps, which are discarded, i.e., the so-called burn-in period. Then, the simulation continues for $n_1=1000$ time steps during which the visits to individual states are counted. All states in which at least $5\%$ of the simulation time $n_1$ is spent are added to the list of pseudo-attractor states.
  

    
    \item[II-1] The simulation of the PBN environment may enter a~fix-point attractor not detected in Step~I. If the simulation gets stuck in a~particular state for $n_2=1000$ steps, the state is added to the list of pseudo-attractor states.
    
    \item[II-2] During training, the simulation of the PBN environment may enter a~multi-state attractor that has not been detected in Step~I. For this reason, a~history of the most recently visited states is kept. When the history buffer reaches the size of $n_3=10000$ items, revisits for each state are counted and states revisited more than 15\% of times are added to the list of pseudo-attractor states. If no such state exists, the history buffer is cleared and the procedure continues for another $n_3$ time steps. The new pseudo-attractor states are added provided no known pseudo-attractor state was reached. Otherwise, the history information is discarded. 
\end{enumerate}

Notice that PASIP allows us to identify the pseudo-attractor states, but does not allow us to assign them to individual pseudo-attractors. Therefore, when training a~DRL agent with the use of pseudo-attractor states, we consider the control strategies between all source-target pairs of pseudo-attractor states. Notice that for large networks where no information on attractors is available, the environment pre-processing phase is important as it provides an~initial pool of pseudo-attractor states for the training.

When identifying a~pseudo-attractor associated with a~PBN attractor, the size of the latter is unknown. Hence, one cannot determine the exact probability threshold in Def.~\ref{def:pseudo-attractor}. PASIP addresses this issue as follows. The $5\%$ identification threshold in Step~I enables the detection of all states of a~complete PBN attractor of size up to $20$ as follows from the following theorem.

\begin{theorem}
\label{th:p-a_size_bounds}
For any PBN attractor $A$, the size of the associated pseudo-attractor $PA$ found by Step~I of the pseudo-attractor identification procedure with $k\%$ identification threshold is exactly upper bounded by
\[
|PA| \leq 
\begin{cases}
\frac{100}{k}-1, & 100 \text{ mod } k = 0 \text{ and } |A| > \frac{100}{k} \\
\floor{\frac{100}{k}}, & \text{otherwise}.
\end{cases}
\]
\end{theorem}

Here we provide a~concise proof of Th.~\ref{th:p-a_size_bounds}. An~elaborated version of the proof can be found in Appendix~\ref{app:pseudo-attractors}.

\ifx\noproof\undefined
\begin{proof}
    Let $S = \frac{100}{k} -1$ if $k \mid 100$ and $|A| = \floor{\frac{100}{k}}$ and assume that we have an~attractor $A$ of size strictly greater than $S$. Then by the pigeonhole principle one of the states has to be visited less then $\frac{100}{S}\% < k\%$ of times. So it will not be fully recovered by the procedure. Hence $|PA| < S + 1$.

    Contrary, if  we have an~attractor $A$ of size $S$, which has uniform distribution then $PA$ will equal exactly $|A|$, so the upper bound for the size of $|PA|$ is at least $S$.
\end{proof}
\fi

In light of Def.~\ref{def:pseudo-attractor} and Th.~\ref{th:uniform}, PASIP can detect as pseudo-attractor states a~complete attractor of size $20$ only if the stationary distribution on the attractor is uniform. If the attractor size is less than 20, PASIP can still detect all attractor states even if the distribution is non-uniform. 
Notice that with decreasing size of the unknown attractor, PASIP allows more and more pronounced deviations from the uniform distribution while preserving the complete attractor detection capability provided the stationary probabilities of all attractor states are above the identification threshold. However, if an~attractor is of size larger than $20$ states, Step~I of PASIP with $5\%$ identification threshold will identify the complete associated pseudo-attractor only if the stationary distribution is non-uniform and the pseudo-attractor will contain only the most frequently revisited states. The maximum possible size of the identified pseudo-attractor in this case is $19$, which follows from Th.~\ref{th:p-a_size_bounds}. 
This is a~desired property of PASIP as it keeps the number of pseudo-attractor states manageable. This has significant positive influence on stabilising DRL agent training.
Our experiments with small networks, i.e., ones for which exact attractors could be computed, revealed that it is beneficial to underestimate the set of attractor states in Step~I of PASIP as the missed states are usually discovered later during the DRL agent training phase.

The environment pre-processing phase provides an~initial set of pseudo-attractor states.
The set is expanded during the DRL agent training phase. Step~II-1 allows to identify plausible fix-point attractors. Step~II-2 enables detection of plausible multi-state attractors. However, here the focus is on smaller attractors than in the case of Step~1: we classify states as pseudo-attractor states if they are revisited at least $15\%$ of time, which 
corresponds to attractors of size 6. This is to restrict the number of spurious pseudo-attractor states in order to stabilise DRL agent training.

\subsection{Exploration probability boost}
\label{sec:expl_prob_boost}
\begin{figure}[ht]
\centering
  \subfloat[Training without EPB.]{
    \includegraphics[width=.481\linewidth]{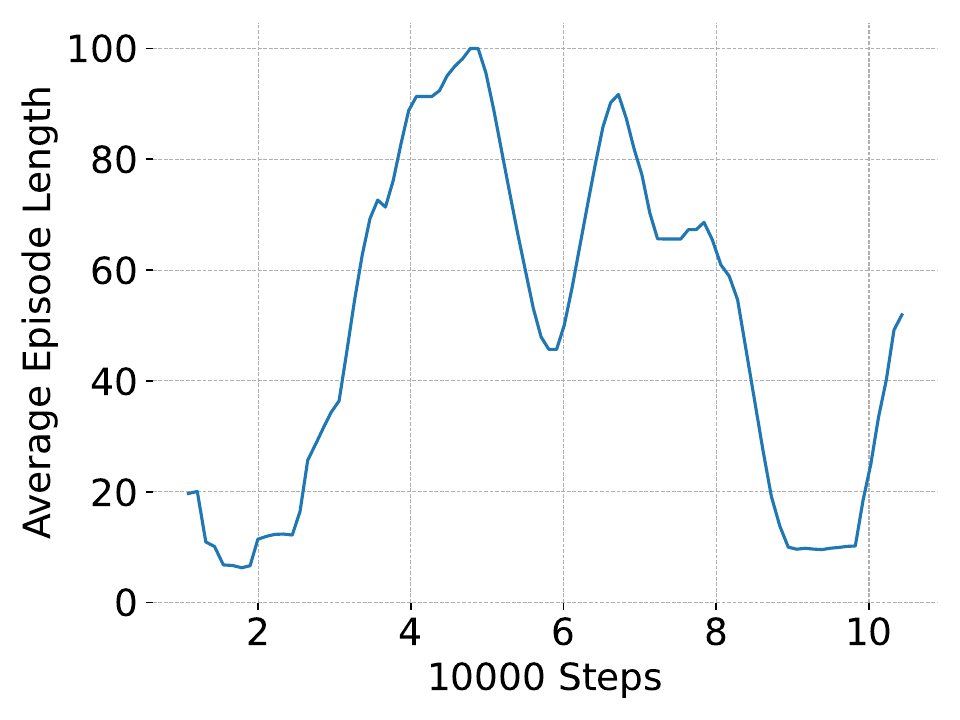}%
    \label{Fig:Data1}}
  \hfill
  \subfloat[Training with EPB.]{
    \includegraphics[width=.481\linewidth]{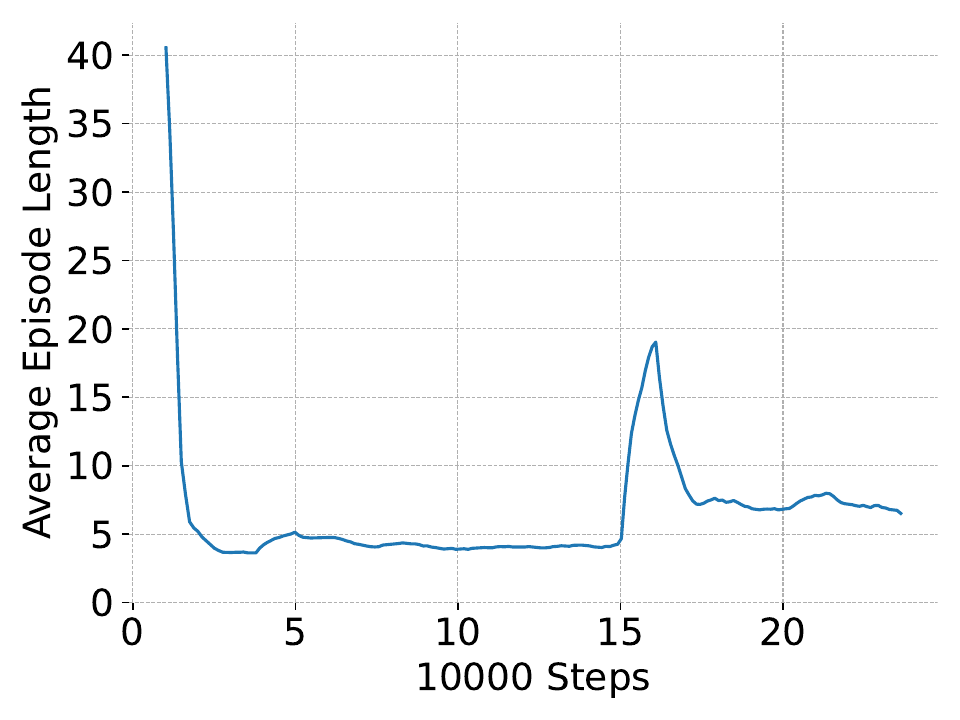}%
    \label{Fig:Data2}}

  \caption{Average episode lengths during DRL agent training run without and with EPB while new pseudo-attractor states are being found causing abrupt episode length increases.}
\end{figure}

We have encountered an~issue related to late discovery of pseudo-attractor states during training. As can be observed in Fig.~\ref{Fig:Data1}, PASIP may detect a~new pseudo-attractor at any point in time which destabilises training: a~new state is detected at around 90000 steps, which causes abrupt, significant increase of the average episode length. We proceed to propose a~remedy to this problem.

To balance exploitation and exploration of the DRL agent during training, we consider the standard $\varepsilon$-greedy policy of DRL which takes random action with the \emph{exploration probability} $\varepsilon$ and with probability $1-\varepsilon$ follows the greedy policy, i.e., it selects the action $a^* = \arg\max_{a \in \mathcal{A}} Q(s, a)$, see Appendix~\ref{app:rl}. We set the initial $\varepsilon$ value to 1 and linearly decrease it to 0.05 over the initial 3000 steps of training.
However, simple combining of the original $\varepsilon$-greedy policy with online identification of pseudo-attractor states gives rise to unstable training. When trying to train the DRL agent for our control problem 
while keeping identifying pseudo-attractor states during training, stability issues mentioned above were observed. To alleviate this negative influence on training, we introduce the \emph{exploration probability boost} (EPB) to the $\varepsilon$-greedy policy. The idea of EPB is to increase the exploration probability $\varepsilon$ after each discovery of a~new pseudo-attractor to $\max(\varepsilon, 0.3)$ if the current value of $\varepsilon$ is less than 0.3. After the increase, the linear decrease to 0.05 follows with the rate of the initial decrease. As revealed by our experiments, this simple technique makes learning much more stable. This is illustrated in Fig.~\ref{Fig:Data2}, where the DRL agent discovers new pseudo-attractor states at around the 150000-th training step, which imposes an~increase in the average episode length. However, the use of the improved $\varepsilon$-greedy policy allows to significantly reduce the magnitude of the increase and results in a~quick return to the previously trained low value of the average episode length as compared to the behaviour in Fig.~\ref{Fig:Data1}.

\section{Implementation}
\label{sec:implementation}
We implement pbn-STAC as a~fork of gym-PBN~\citep{sota-gym}, an~environment for modelling of PBNs, and pbn-rl~\citep{sota-pbn-rl-git}, a~suite of DRL experiments for a~different PBN control problem formulated in~\citep{MCSW22}.
In pbn-STAC, we have adapted the original code of gym-PBN and pbn-rl to our formulation of the PBN control problem, i.e., the source-target attractor control. First, we extend gym-PBN by adding the asynchronous PBN environment to it. Second, to allow for simultaneous perturbation of a~combination of genes within a~DRL action, we replace the original DDQN architecture with the Branching Dueling Q-Network (BDQ) architecture~\citep{bdq}, which, contrary to DDQN, scales linearly with the dimension of the action space. The architecture of our BDQ network is described in Appendix~\ref{app:bdq}.
Third, we implement the pseudo-attractor states identification procedure and the exploration probability boost technique. Finally, the framework takes as input a~source-target pair of attractors or pseudo-attractor states. In the case of a~multi-state target attractor, a~training episode is regarded as successful if any of the target attractor states is reached. For a~multi-state source attractor, we uniformly sample one of its states and set it as the initial state for an~episode. 
Our DRL-based framework for the source-target attractor control is made available via the dedicated pbn-STAC GitHub repository~\citep{git_pbn_rl}.

\section{Experiments}
\label{sec:experiments}

\subsection{BN and PBN models of GRNs}

\noindent\textbf{Melanoma models.} We infer BN and PBN environments of various sizes for the melanoma GRN using the gene expression data provided by Bittner \emph{et al.} in~\citep{bittner}. This is a~well-known dataset on melanoma, which is extensively studied in the literature, see, e.g.,~\citep{MCSW22,Du2023}.
We infer the BN/PBN structures with the method implemented in gym-PBN, see~\citep{MCSW22} for details. It is based on the coefficient of determination (COD), which is a~measure of how well the dependent variable can be predicted by a~model, which in the case of~\citep{MCSW22} is a~perceptron. All our BN and PBN models are built around the following seven core genes:
reticulon 1;
hydroxyacyl-Coenzyme A dehydrogenas;
ESTs;
pirin;
melan-A;
wingless-type MMTV integration site family, member 5A; and
S100 calcium-binding protein, beta (neural).
We refer to the Suppl. Materials for full lists of genes of individual models.

\noindent\textbf{Case study of {\it B.~bronchiseptica}.}
We test pbn-STAC on an~existing model of a~real biological system, i.e., the network of immune response against infection with the respiratory bacterium \emph{B.~bronchiseptica}, which was originally proposed and verified against empirical findings in~\citep{BB}. The computational model, denoted IRBB-33, is an~asynchronous BN consisting of 33 genes.

\subsection{Performance evaluation methodology}
We evaluate the performance of pbn-STAC in solving the control problem formulated in Sec.~\ref{sec:control_problem} on both BN and PBN models of melanoma of various sizes, i.e., incorporating 7, 10, and 30 genes. Moreover, we consider IRBB-33, the 33-genes BN model. The dynamics under the asynchronous update mode is considered for all models.
The evaluation consists of the DRL agent interacting with the environment by taking actions, where an~action consists of flipping the values of a~particular subset of genes in a~(pseudo-)attractor state. We recover a~control strategy for a~given source-target pair learned by a~trained DRL agent by initialising the BN/PBN environment with the source and target and letting it run with $\varepsilon$ set to zero while applying the actions suggested by the DRL agent.
To evaluate the performance of pbn-STAC on a~particular BN/PBN model, we recover control strategies for all possible ordered source-target pairs of the model's (pseudo-)attractor states.
For all the BN models of melanoma and IRBB-33, we are able to compute all their exact attractors and, with exception of BN-30, the optimal control strategies for all pairs of attractors using the attractor-based sequential instantaneous source-target control (ASI) of the CABEAN software tool~\citep{SP21}. We use the information on the exact attractors and optimal control strategies for BN models as ground truth for the evaluation of pbn-STAC.

For PBN-7 and PBN-10, we compute the attractors with the NetworkX package~\citep{NetworkX}, which facilitates the analysis of complex networks in Python. Unfortunately, due to very large memory requirements, we are unable to obtain the attractors of the 30-genes PBN model of melanoma with this approach, so we consider pseudo-attractor states. The optimal-length control strategies are obtained for PBN-7 and PBN-10 models by exhaustive search.

Notice that due to the nondeterministic nature of our environments, i.e., the asynchronous update mode, the results may vary between runs. Thus, for each source-target pair, we repeat the run 10 times. Therefore, for each model the total number of recovered pbn-STAC control strategies is 10x the number of distinct source-target pairs. For each recovered control strategy, we count its length and record the information on being successful, i.e., whether the target state is reached with a~strategy not exceeding maximum allowed length. 
The reported average lengths are computed by considering the successful runs only. 

All reference strategies found by CABEAN or exhaustive search are deterministic and successful, thus the averaging is performed over the number of distinct source-target pairs without run repetitions. CABEAN provides us with multiple strategies, referred to as `paths' for BN models. All have the same minimal total number of gene perturbations, but they are divided into interventions in different ways. For each individual source-target pair, we consider the length of the shortest strategy possible which perturbs at most 3 genes per intervention.
For the exhaustive search approach in the case of PBN models, only strategies that perturb at most 3 genes in an~intervention are considered and the length of the shortest one is considered for a~particular source-target pair. For both approaches, we compute the average over the lengths for all source-target pairs.

\section{Results}
\label{sec:results}

\subsection{Identification of pseudo-attractor states}

We evaluate the performance of PASIP proposed in Sec.~\ref{sec:pas-procedure}. For this purpose, we run pbn-STAC with PASIP on the considered BN/PBN models. We present the obtained results in Tab.~\ref{tab:pas-procedure_res}. For each model, except the melanoma PBN-30 for which the exact attractors could not be obtained, we provide the information on the number of exact attractors, the total number of exact attractor states, and the total number of identified pseudo-attractor states with PASIP. We asses PASIP using the F1 score defined as
$2\,\textrm{TP}/(2\textrm{TP}+\textrm{FP}+\textrm{FN})$,
where $\textrm{TP}$ is the number of true positives, i.e., the number of pseudo-attractor states that indeed are attractor states, $\textrm{FP}$ is the number of false positives, i.e., the number of states identified as pseudo-attractor states which are not part of any of the network's attractor, and $\textrm{FN}$ is the number of false negatives, i.e., the number of states not identified as pseudo-attractor states but which are the network's attractor states. 
Our procedure achieves the maximal value of the F1 score, i.e., 100\%, in all test cases in which the exact attractors are known, which means that PASIP does not introduce any FPs and FN, resulting in perfect precision, i.e., $\textrm{TP}/(\textrm{TP}+\textrm{FP})$, and recall, i.e., $\textrm{TP}/(\textrm{TP}+\textrm{FN})$. In the case of PBN-30, all pseudo-attractor states were verified to indeed be fixed point attractors, resulting in zero FPs. However, since the number of all exact attractor states is unknown, 
the F1 score cannot be determined (indicated with N/A in Tab.~\ref{tab:pas-procedure_res}). Nevertheless, the precision in this case is 100\%.

We experimented with various settings of the PASIP hyper-parameters and found that the results do not vary significantly (data not shown), indicating that PASIP is rather robust with respect to the setting of the hyper-parameters. 
In summary, the presented results show that PASIP is reliable.

\begin{table}
        \begin{tabular}{|c|c|c|c|c|}
             \hline
             Model & \textnumero\,Attr. & \textnumero\,Attr. states & \textnumero\,PA-states & F1\\
             \hline\hline
              BN-7 & 6 & 6 & 6 & 100\%\\ 
             \hline
             BN-10 & 26 & 26 & 26 & 100\%\\
             \hline
             BN-30 & 148 & 148 & 148 & 100\%\\
             \hline
            IRBB-33 & 3 & 3 & 3 & 100\%\\
             \hline\hline
            PBN-7 & 4 & 4 & 4 & 100\%\\ 
             \hline
             PBN-10 & 6 & 6 & 6 & 100\%\\
             \hline
             PBN-30 & unknown & unknown & 32 & N/A\\
             \hline
        \end{tabular}
    \caption{Performance of PASIP measured by the F1 score. Numbers of exact attractors and all their states versus pseudo-attractor states (PA-states) identified by PASIP are shown. We were unable to compute the exact attractors for the PBN-30 model, which is indicated with `unknown'. Attr. is short for attractor.}
    \label{tab:pas-procedure_res}     
\end{table}


\begin{table}
\begin{tabular}{|l|c|c|c|c|} 
         \hline
         Model & \textnumero\,S-T pairs & Opt. strategy & pbn-STAC & Failure rate\\  
         \hline\hline
         BN-7 & 30 & 1.0 & 2.64 & 0.0\%\\ 
         \hline
         BN-10 & 650 &1.1 & 2.04 & 4.0\%\\
         \hline
         BN-30 & 21756 & unknown & 3.08 & 5.7\%\\
         \hline
         IRBB-33 & 6 & 1.0 & 1.33 & 0.0\%\\
         \hline\hline
         PBN-7 & 12 & 1.1  & 1.63 & 0.0\%\\ 
         \hline
         PBN-10 & 30 & 1.2 & 1.53 & 0.0\%\\
         \hline
         PBN-30 & 992 & unknown & 8.02 (2.37) & 2.8\%\\
         \hline
    \end{tabular}
    \caption{Average lengths of pbn-STAC control strategies and optimal ones (Opt. stategy) obtained with CABEAN (BNs) or exhaustive search (PBNs) for all source-target pairs (S-T pairs) of (pseudo-)attractor states of individual models. For the PBN-30 model, pseudo-attractor states were used. The impossibility to obtain optimal strategies is indicated with `unknown'.}
\label{tab:control_res}
\end{table}

\subsection{Control of melanoma models}
We evaluate the ability of pbn-STAC to solve the control problem by comparing the obtained results for BN models with the optimal ASI control strategies computed with CABEAN. We also run the pbn-STAC on the three PBN models of melanoma. For PBN-30, we are not able to compute the set of exact attractors, but we identify 32 pseudo-attractor states with PASIP.

\begin{figure}[!ht]
\centering
   \subfloat[BN-7]{
     \includegraphics[width=.5\linewidth]{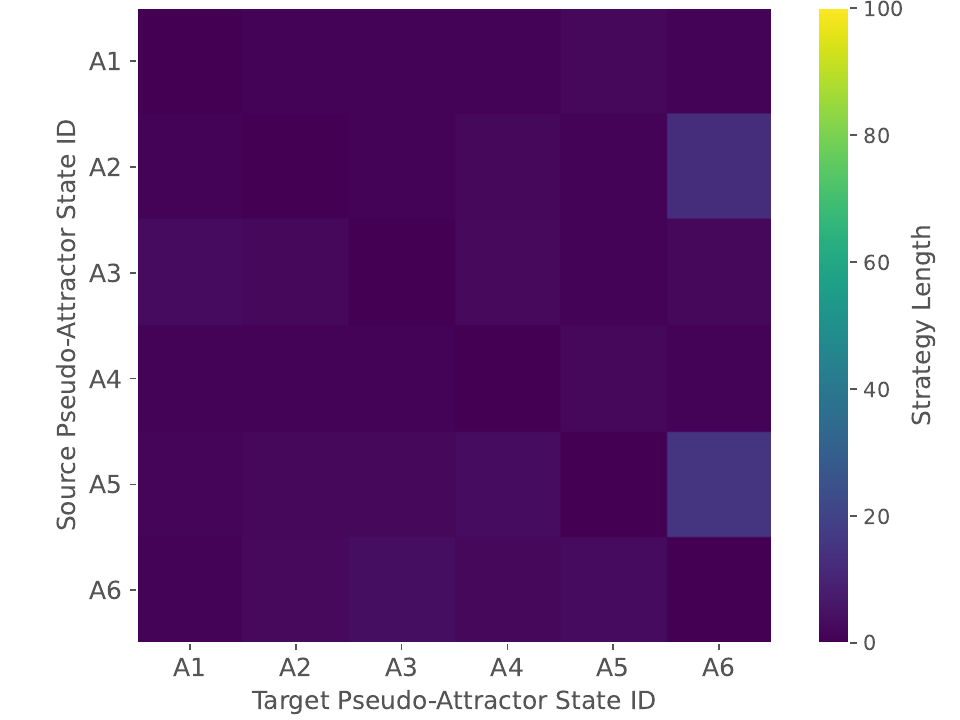}\label{fig:bn7:heat}}
    \subfloat[BN-10]{
     \includegraphics[width=.5\linewidth]{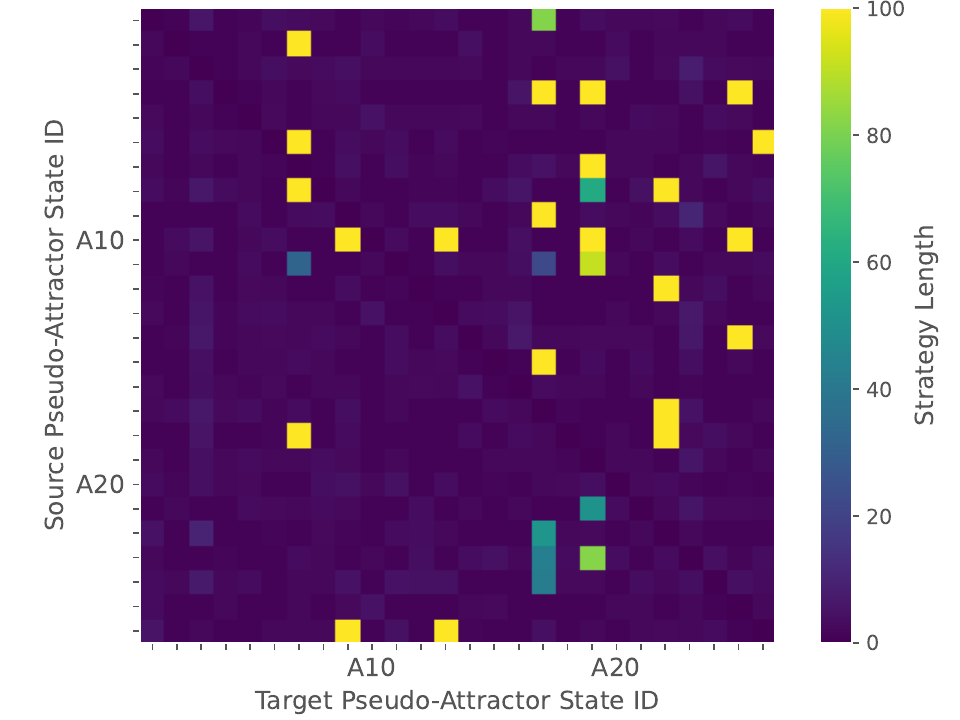}\label{fig:bn10:heat}}
 \medskip
 
     \subfloat[BN-30]{
     \includegraphics[width=.5\linewidth]{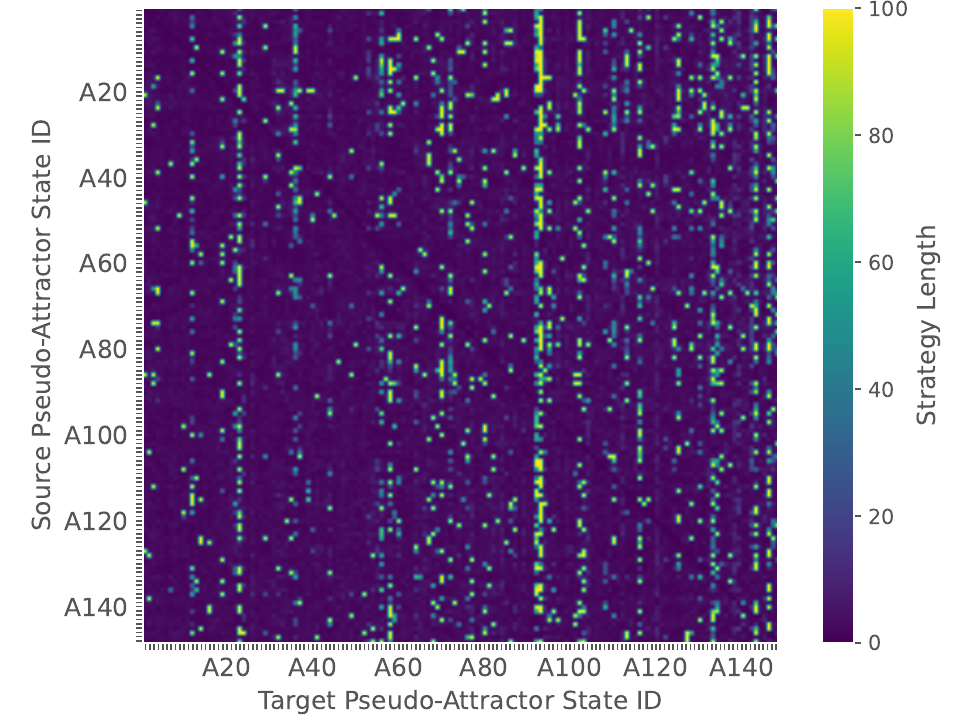}\label{fig:bn30:heat}}
     \subfloat[IRBB-33]{
     \includegraphics[width=.5\linewidth]{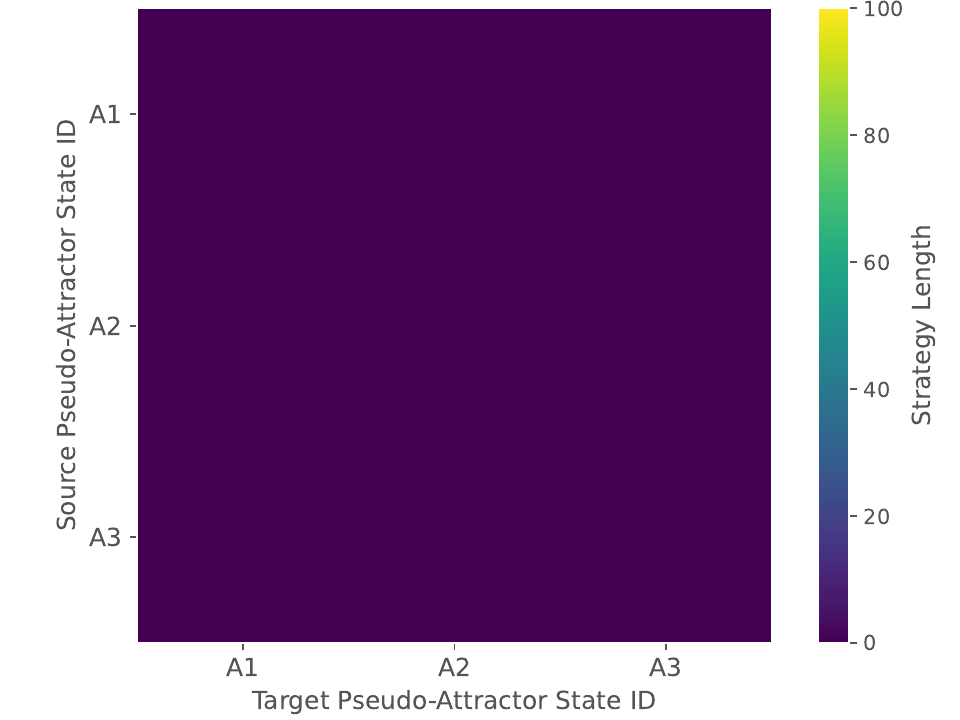}\label{fig:bn33:heat}}
     
  \subfloat[PBN-7]{
    \includegraphics[width=.5\linewidth]{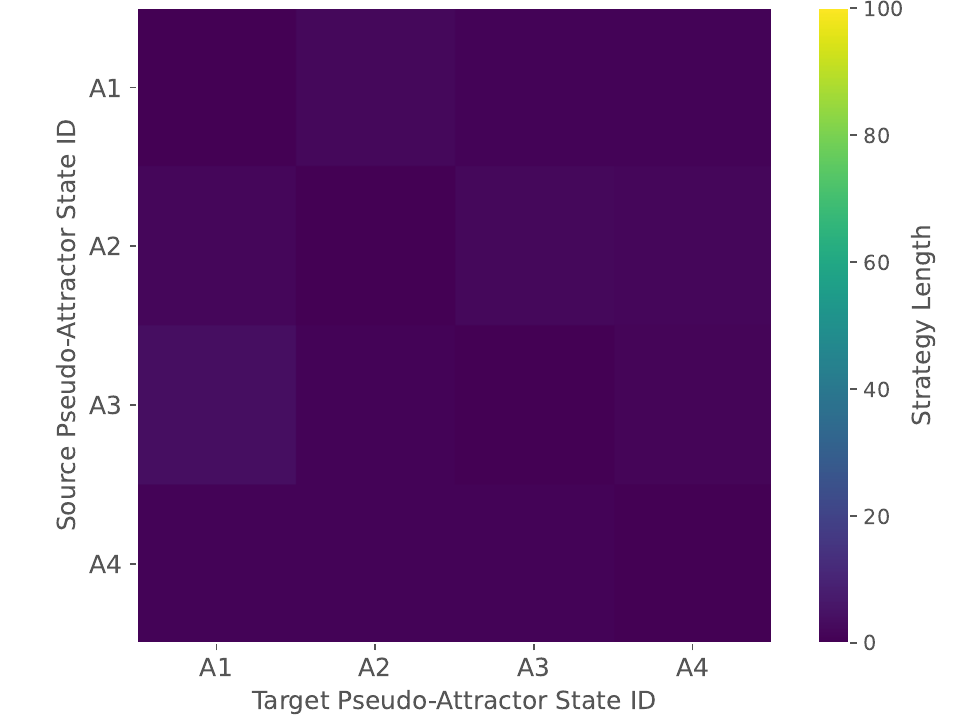}\label{fig:pbn7:heat}}
  \subfloat[PBN-10]{
     \includegraphics[width=.5\linewidth]{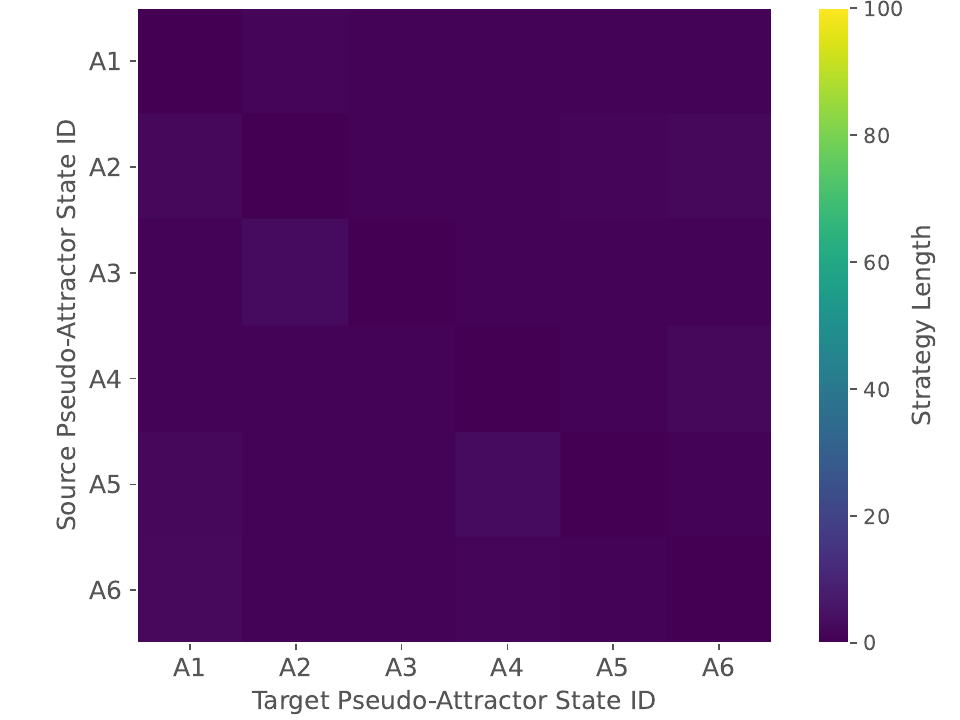}\label{fig:pbn10:heat}}
     
  \subfloat[PBN-30]{
  \includegraphics[width=.5\linewidth]{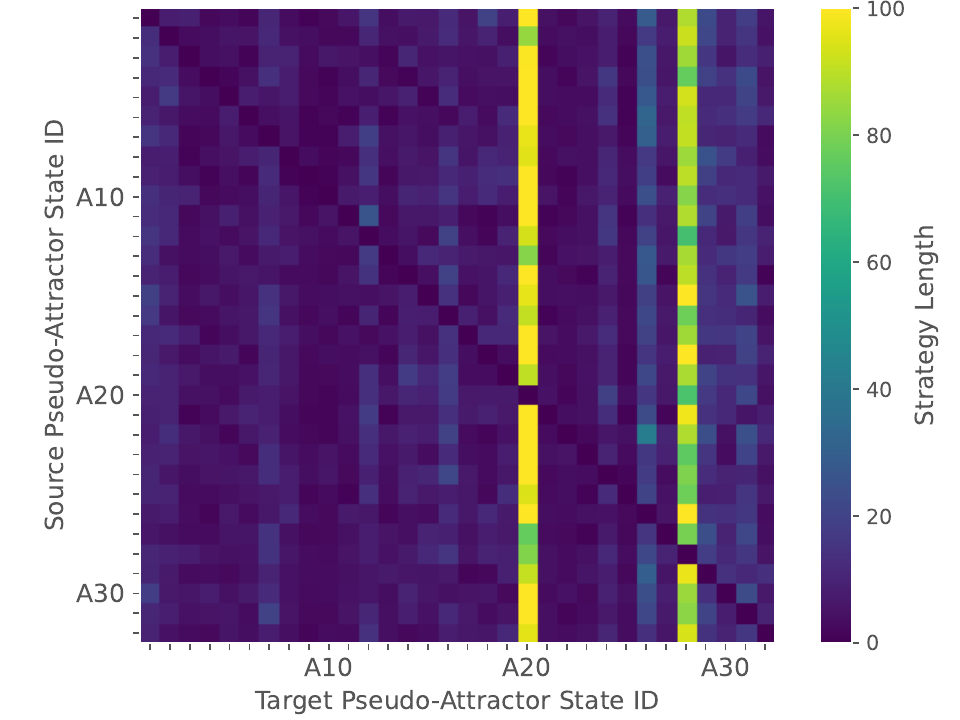}\label{fig:pbn30:heat}}   
  \caption{Heatmaps of strategy lengths for individual source-target (pseudo-)attractor states averaged over 10 runs.}
  \label{fig:model_tester_out_heat}
\end{figure}

As can be seen in Tab.~\ref{tab:control_res}, the successful strategies obtained with pbn-STAC for all models on average tend to be comparable in length to the optimal ones. The failure rate in Tab.~\ref{tab:control_res} indicates the percentage of pbn-STAC strategies that failed to reach the target state. To investigate the slight increases in average lengths, we compute heatmaps of averaged strategy lengths over 10 runs for each individual source-target (pseudo-)attractor states pair shown in Fig.~\ref{fig:model_tester_out_heat} (see Fig.~\ref{fig:model_tester_out_heat_bn:appendix}
\& \ref{fig:model_tester_out_heat_pbn:appendix} in Appendix~\ref{app:results} for larger versions). The plots reveal that in the majority of cases pbn-STAC is capable of identifying the optimal or sub-optimal strategies. However, in some cases longer strategies are also returned resulting in long strategy length distribution tails, showed in detail in histograms of all successful control strategy lengths provided in Fig.~\ref{fig:BN_histograms:appendix} \& \ref{fig:PBN_histograms:appendix} of Appendix~\ref{app:results}.

Although the optimal control strategies for PBN-30 could not be obtained, the average strategy length of 8.02 computed with pbn-STAC is significantly larger than in the other cases. As can be seen in Fig.~\ref{fig:pbn30:heat}, two target pseudo-attractor states, i.e., A20 and A27, are very hard to reach, which could indicate that their basins of attraction are very small. When all source-target pairs with A20 or A27 as the target are excluded from analysis, the average length drops to 2.37, which is reported in brackets in Tab.~\ref{tab:control_res}.


 
     

\subsection{Control of the IRBB-33 model}

\begin{figure}[ht]
  \centering
  \subfloat[Mixed reward of Sec.~\ref{sec:algorithm}]{
    \includegraphics[width=.48\linewidth]{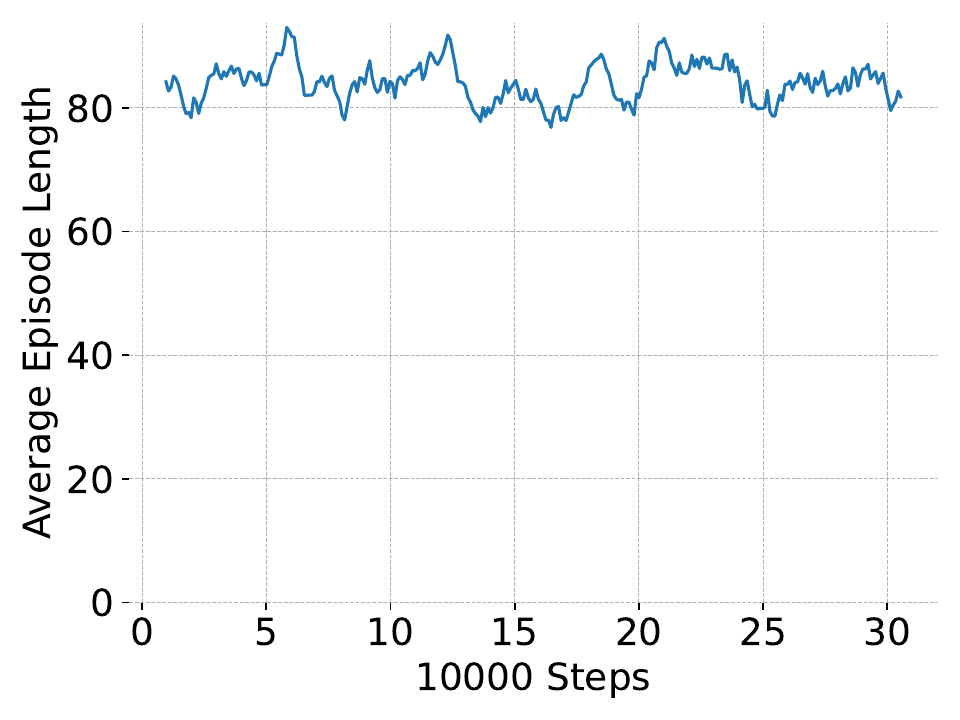}\label{fig:bn33_data1}}
  \subfloat[Improved reward]{
     \includegraphics[width=.48\linewidth]{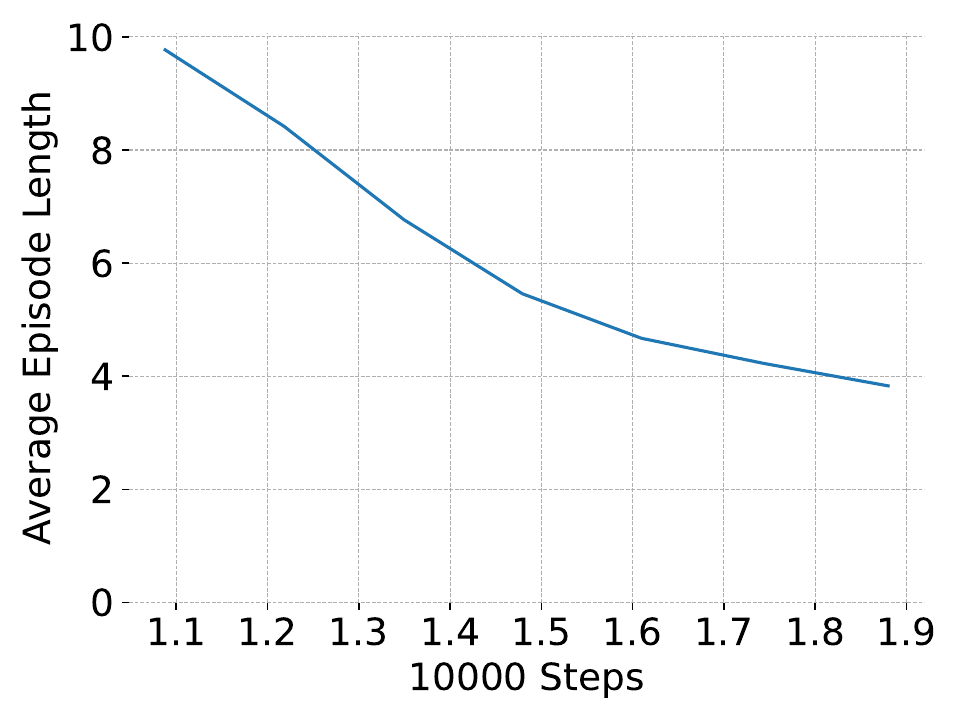}\label{fig:bn33_data2}}
  \caption{Training of the DRL agent on the IRBB-33 environment with different reward schemes.}
\end{figure}

In the case of the IRBB-33 network, we have to modify the reward scheme. As can be seen in Fig.~\ref{fig:bn33_data1}, the reward scheme introduced in Sec.~\ref{sec:algorithm}, referred to as the \emph{mixed reward}, does not lead to any improvement of the average episode length during training over 200\,000 training steps.
After trying different reward schemes for this network (data not shown), we found that the reward function $R_a(s, s') =  -|a| + 100 * (\mathbbm{1}_{TA}(s') - 1)$ improves the training of the DRL agent significantly, as can be seen in Fig.~\ref{fig:bn33_data2}, where the convergence is achieved. 
The new reward scheme is based on the same idea of penalising for many actions and awarding the reaching of the target state, but is shifted to be negative. Reward function having constant sign can sometimes interplay with the rest of the hyperparameters of DRL making learning more stable. We just mention here, that the improved reward scheme is robust in the sense that it can be used instead of the previously introduced reward in Sec.~\ref{sec:algorithm} for all the BN/PBN melanoma models considered in this study without deteriorating the results.

As presented in Tab.~\ref{tab:control_res}, the average control strategy length obtained with pbn-STAC employing the new reward scheme is $1.58$, which is almost optimal. The detailed heatmap is provided in Fig.~\ref{fig:bn33:heat}.

\section{Discussion}
\label{sec:conclusions}

In this study, we formulate a~novel control problem for the BN and PBN frameworks under the asynchronous update mode that corresponds with the problem of identifying cellular reprogramming strategies. We develop and implemented a~computational framework based on DRL, i.e., pbn-STAC, that solves the control problem. Given source and target (pseudo-)attractor states, pbn-STAC finds proper control strategies that drive the network from the source to the target by intervening only in intermediate attractor states that correspond to phenotypical cellular states that can be observed in wet-lab practice. Since identifying attractors of large BNs/PBNs is a~challenging problem by itself and we consider our framework as a~contribution towards developing scalable control methods for large networks, we introduce the notion of a~pseudo-attractor and develop a~procedure that identifies pseudo-attractor states during DRL agent training, i.e., PASIP. We evaluate the performance of PASIP and pbn-STAC on a~number of biological networks of various sizes. We compare pbn-STAC predictions with the exact, optimal solutions wherever possible.

The obtained results show the potential of pbn-STAC in terms of effectiveness and at the same time reveal some points for improvement. First, addressing the problem of long tails of the distributions of the lengths of strategies identified by pbn-STAC would allow us to  improve the performance of pbn-STAC. Second, for large models, the set of pseudo-attractor states may take a~long time to stabilise. The computations of pseudo-attractor states can be parallelised in a~rather straightforward way to speed up the detection. Third, as described in the literature, finding fixed-point attractors is somewhat easier than finding multi-state attractors. Therefore, PASIP could be modified to compute all fixed-point attractors with existing state-of-the-art techniques in Step~I, rendering Step~II-1 unnecessary. Finally, we perceive our framework as rather straightforwardly adaptable to other types of PBNs, such as PBNs with perturbations, or Probabilistic Boolean Control Networks. We consider these developments and the evaluations of pbn-STAC on models of larger sizes as future work.


\bibliographystyle{abbrv}
\bibliography{mybib}

%

\clearpage

\appendix

\setcounter{theorem}{0}

\section{Boolean and probabilistic Boolean networks}
\label{app:BN_PBN}

A~Boolean network consists of nodes that can be in one of two states, and Boolean functions describing how the individual nodes interact with each other.

\begin{definition}[Boolean network]
A~\emph{Boolean network} (BN) is defined as a~pair $(V, F)$, where $V = \{x_1, x_2, \ldots, x_n\}$ is a~set of binary-valued nodes (also referred to as genes) and $F = \{f_1, f_2, \ldots, f_n\}$ is a~set of Boolean predictor functions, where $f_i(x_{i_1},x_{i_2}, \ldots, x_{i_k})$ defines the value of node $x_i$ depending on the values of the $k \leq n$ parent nodes $x_{i_1},x_{i_2},\ldots, x_{i_k}$ with $i_j \in [1..n]$ for $j\leq k$.
\end{definition}

Probabilistic Boolean networks are an~extension of the formalism of BNs. Formally, probabilistic Boolean networks are defined as follows:

\begin{definition}[Probabistic Boolean network]
A~Probabistic Boolean network (PBN) is a~discrete dynamical system defined as a~pair $(V, \mathcal{F})$, where $V = \{x_1, x_2, \ldots, x_n\}$ is a~set of binary-valued nodes (also referred to as genes) and $\mathcal{F} = (F_1, F_2,\ldots, F_n)$ is a~list of sets. Each node $x_i \in V$ has associated a~set $F_i \in \mathcal{F}$ of predictor functions, i.e., $F_i = \{f^i_1, f^i_2, \ldots, f^i_{l(i)}\}$, where $l(i)$ is the number of predictor functions of node $x_i$. Each $f^i_j \in F_i$ is a~Boolean function defined with respect to a~subset of $V$ denoted $\textrm{Pa}(f^i_j)$ and referred to as parent nodes of $f^i_j$. For each $x_i \in V$, there is a~probability distribution $\mathbf{c}^i = (c^i_1, c^i_2, \ldots, c^i_{l(i)})$ on $F_i$, where each predictor function $f^i_j \in F_i$ has an~associated selection probability denoted $c^i_j$; it holds that $\sum_{j=1}^{l(i)}c^i_j=1$.
\end{definition}

\begin{example}
\label{running}
We consider a~PBN of 4 genes $V=\{x_0, x_1, x_2, x_3\}$ regulated in accordance with the following Boolean functions:

\begin{center}
\begin{tabular}{ c }
$f_0^1(x_0) = x_0$  \\
$f_0^2(x_0, x_1, x_2, x_3) = x_0 \& \neg (x_0 \& \neg x_1 \& \neg x_2 \& x_3)$\\
 $f_1^1(x_0, x_1) = \neg x_0 \& x_1$\\
 $f_1^2(x_0, x_1, x_2, x_3) = \neg x_0 \& (x_1 | (x_2 \& x_3))$ \\
 $f_2^1(x_0, x_1, x_2, x_3) = \neg x_0 \& (x_1 \& x_2 \& x_3)$\\
 $f_2^2(x_0, x_1, x_2, x_3) = x_0 \& (\neg x_1 \& \neg x_2 \& \neg x_3)$ \\
 $f_3^1(x_0, x_1, x_2, x_3) = \neg x_0 \& (x_1 | x_2 | x_3)$\\
 $f_3^2(x_0, x_1, x_2, x_3) = \neg x_0 \& (x_1 | x_2 | x_3)$ 
\end{tabular}
\end{center}
Under the asynchronous update mode, the dynamics of the PBN can be represented with the state transition graph (STG) depicted in Fig.~\ref{fig:stg}. The PBN has two fixed-point attractors, namely $\{(0,0,0,0)\}$ and $\{(0, 1, 0,1)\}$ and one multi-state attractor, namely $\{(1,0,0,0), (1,0,1,0)\}$.
\end{example}

\begin{figure}[ht]
\centering
\scalebox{1}{%
\begin{tikzpicture}[x=6cm,y=4cm] 
  \tikzset{     
    e4c node/.style={circle,draw,minimum size=0.75cm,inner sep=0}, 
    e4c edge/.style={sloped,above,font=\footnotesize}
  }
  \node[e4c node] (1) at (0.56, 0.9) {(0,0,0,1)}; 
  \node[e4c node] (2) at (0.44, 0.58) {(0,0,1,0)}; 
  \node[e4c node] (3) at (0.89, 0.72) {(0,0,1,1)}; 
  \node[e4c node] (4) at (0.3   , 0.03) {(0,1,0,0)}; 
  \node[fill=blue!30][e4c node] (5) at (0.7, 0.44) {(0,1,0,1)}; 
  \node[fill=blue!30][e4c node] (8) at (0.2, 1.54) {(1,0,0,0)}; 
  \node[e4c node] (7) at (1.00, 0.33) {(0,1,1,1)}; 
  \node[e4c node] (6) at (0.71, 0.03) {(0,1,1,0)}; 
  \node[e4c node] (9) at (0.56, 1.25) {(1,0,0,0)}; 
  \node[fill=blue!30][e4c node] (10) at (0.55, 2.05) {(1,0,1,0)}; 
  \node[e4c node] (11) at (0.88, 1.60) {(1,0,1,1)}; 
  \node[e4c node] (12) at (0.22, 2.1) {(1,1,0,0)}; 
  \node[e4c node] (13) at (0.60, 1.65) {(1,1,0,1)}; 
  \node[e4c node] (14) at (0.54, 2.5) {(1,1,1,0)}; 
  \node[e4c node] (15) at (0.91, 2.00) {(1,1,1,1)}; 
  \node[fill=blue!30][e4c node] (16) at (0.20, 0.41) {(0,0,0,0)};

  \path[->,draw,thick]
  (2) edge[e4c edge]  (3)
  (1) edge[e4c edge]  (2)
  (1) edge[e4c edge]  (5)
  (2) edge[e4c edge ]  (16)
  (4) edge[e4c edge]  (5)
  (4) edge[e4c edge]  (16)
  (3) edge[e4c edge]  (1)
  (3) edge[e4c edge]  (7)
  (15) edge[e4c edge]  (11)
  (15) edge[e4c edge]  (13)
  (15) edge[e4c edge]  (14)
  (14) edge[e4c edge]  (10)
  (14) edge[e4c edge]  (12)
  (13) edge[e4c edge]  (9)
  (13) edge[e4c edge]  (12)
  (12) edge[e4c edge]  (8)
  (12) edge[e4c edge]  (10)
  (11) edge[e4c edge]  (9)
  (11) edge[e4c edge]  (10)
  (10) edge[e4c edge, bend left=20]  (8)
  (9) edge[e4c edge]  (1)
  (6) edge[e4c edge]  (7)
  (6) edge[e4c edge]  (5)
  (6) edge[e4c edge]  (2)
  (7) edge[e4c edge]  (5)
  (9) edge[e4c edge]  (8)
  (8) edge[e4c edge, bend left=20]  (10)
  ;
\end{tikzpicture}
}
\caption{STG of the PBN defined in Example~\ref{running} under the asynchronous update mode. Shaded states are the attractor states of the three attractors, i.e., two fixed-point attractors $A_1=\{(0,0,0,0)\}$ and $A_2=\{(0, 1, 0,1)\}$, and one multi-state attractor $A_3 = \{(1,0,0,0), (1,0,1,0)\}$.}
\label{fig:stg}
\end{figure}

\begin{example}\label{ex:control}
    The PBN from Example~\ref{running} may be controlled from state $(1,0,1,0)$ to $(0,0,0,0)$ by intervening on $x_1$ and allowing the PBN to evolve in accordance with its original dynamics:
    $$(1,0,1,0) \xrightarrow{i=1} (0, 0, 1, 0) \xrightarrow{\text{evolution}} (0, 0, 0, 0).$$

    However, the evolution is non-deterministic and the PBN may evolve to another attractor, see Fig.~\ref{fig:stg}: 
    $$(0, 0, 1, 0) \xrightarrow{\text{evolution}} (0, 1, 0, 1).$$

    The only way to be sure to move to $(0,0,0,0)$ is to flip genes $\{x_1, x_3\}$ either simultaneously, which gives a~strategy of length one, or one-by-one, which gives a~strategy of length two, i.e., $[\{x_3\}, \{x_1\}]$.
    
\end{example}

\ifx\nopa\undefined
\section{Pseudo-attractors}
\label{app:pseudo-attractors}

\begin{theorem}
\label{th:existence:appendix}
Let $A$ be an~attractor of a~PBN. Then there exists a~pseudo-attractor $PA \subseteq A$ such that $|PA| \geq 1$.     
\end{theorem}

\begin{proof}
Let $n$ be the size of the attractor $A$, i.e., $n:=|A|$. Since the underlying Markov chain of the PBN restricted to $A$ is irreducible and positive recurrent, it has a~unique stationary distribution, which we denote $\mathbb{P}_A$. We proceed to show that there exists at least one state $s' \in A$ such that $\mathbb{P}_A(s') \geq \frac{1}{n}$. For this, let us assume that no such state exists. Then, we have that
$$
\sum_{s \in A}\mathbb{P}_A(s) < \sum_{s\in A}\frac{1}{n} = n\cdot\frac{1}{n} = 1.
$$
The left-hand side of the above inequality is strictly less than 1 and hence $\mathbb{P}_A$ is not a~probability distribution on $A$, which leads to a~contradiction. In consequence, $|PA| \geq 1$.
\end{proof}

\begin{theorem}
\label{th:uniform:appendix}
In the case of the~uniform stationary distribution on an~attractor, the associated pseudo-attractor is equal to the attractor:
Let $A$ be an~attractor of a~PBN such that the unique stationary distribution of the underlying Markov chain of the PBN restricted to $A$ is uniform. Then, for the pseudo-attractor $PA$ associated with $A$ it holds that $PA=A$. 
\end{theorem}

\begin{proof}
Let $n$ be the size of the attractor $A$. By the assumption of uniformity of the stationary distribution $\mathbb{P}_A$ it holds that $\mathbb{P}_A(s) = \frac{1}{n}$ for each $s \in A$. Since the pseudo-attractor $PA$ is the maximal subset of $A$ such that $\mathbb{P}_A(s) \geq \frac{1}{n}$ for each $s \in A$, it follows that $PA=A$.
\end{proof}
\fi

\section{Size of a~pseudo-attractor}

Herein, we provide an~elaborated version of the proof of Th.~\ref{th:p-a_size_bounds}. 

\begin{theorem}
For any PBN attractor $A$, the size of the associated pseudo-attractor $PA$ found by Step~I of the pseudo-attractor identification procedure with $k\%$ identification threshold is exactly upper bounded by
\[
|PA| \leq 
\begin{cases}
\frac{100}{k}-1, & 100 \text{ mod } k = 0 \text{ and } |A| > \frac{100}{k} \\
\floor{\frac{100}{k}}, & \text{otherwise}.
\end{cases}
\]
\end{theorem}

\begin{proof}
Since all states of $A$ are positive recurrent in the PBN underlying Markov chain restricted to $A$, it holds that $\mathbb{P}_A(s) > 0$ for all $s \in A$. We start with the first case, where $100 \text{ mod } k = 0$ and $|A| > \frac{100}{k}$. A~state $s$ is identified as a~pseudo-attractor state by Step~I if $\mathbb{P}_A(s) \geq \frac{k}{100}$. A~pseudo-attractor associated with $A$ will be of maximum size whenever the stationary distribution maximises the number of states with $\frac{k}{100}$ probability. The maximum must be less than $\frac{100}{k}$. To show this, let $M$ denote the subset of states of $A$ with stationary distribution probability of $\frac{k}{100}$ and let assume that $|M| \geq \frac{100}{k}$. Then, $\sum_{s \in A}\mathbb{P}_A(s) =  \sum_{s \in M}\mathbb{P}_A(s) + \sum_{s \in A\setminus M} \mathbb{P}_A(s) \geq \frac{k}{100} \cdot \frac{100}{k} + \sum_{s \in A\setminus M} \mathbb{P}_A(s) > 1$, where the last inequality follows from the fact that all the states in $A\setminus M$ have non-zero stationary probabilities. Therefore, $\mathbb{P}_A$ cannot be a~probability distribution. Hence, $|M| \leq \frac{100}{k} - 1$. The upper bound is reached, for example, by any~stationary distribution $\mathbb{P}_A$ such that the for the $\frac{100}{k} - 1$ states in $M$ the probabilities are equal to $\frac{k}{100}$ and $\sum_{s \in A\setminus M} \mathbb{P}_A(s) = \frac{k}{100}$. This concludes the first case. 

In the other case, let $\epsilon := \frac{100}{k} - \floor{\frac{100}{k}}$. If $|A| \leq \floor{\frac{100}{k}}$, then the inequality holds since $PA \subseteq A$ by definition. The upper bound is reached if $|A| = \floor{\frac{100}{k}}$ and, for example, the stationary distribution probabilities for $|A|-1$ states in $A$ are equal to $\frac{k}{100}$ and for the remaining state the probability is $\frac{k}{100}\cdot\left(1+\epsilon\right)$. If $|A| > \floor{\frac{100}{k}}$ in the second case, then $\epsilon > 0$, i.e., $\epsilon \in (0,1)$. The maximum possible number of states of $A$ with stationary probability equal $\frac{k}{100}$ must be less than $\floor{\frac{100}{k}} + 1$. To see this, let $M$ be the set of such states of $A$ and let assume that $|M| \geq \floor{\frac{100}{k}} + 1$. Then, $\sum_{s \in A}\mathbb{P}_A(s) > \frac{k}{100}\cdot \left(\floor{\frac{100}{k}} + 1\right) > \frac{k}{100}\cdot\left(\floor{\frac{100}{k}} + \epsilon\right) = 1$. This contradicts the fact that $\mathbb{P}_A$ is a~probability distribution on $A$. Finally, we proceed to show that the upper bound can be reached also in this scenario. This is the case, for example, for a~stationary probability distribution $\mathbb{P}_A$ on $A$ where $\floor{\frac{100}{k}}$ states of set $M$ have probabilities $\frac{k}{100}$ and all the states in $A\setminus M$ have non-zero probabilities such that $\sum_{s \in A\setminus M} \mathbb{P}_A(s) = \epsilon$. This is possible since $\epsilon > 0$.
\end{proof}

\begin{example}
For the network defined in the running Example~\ref{running}, the true attractor states are $(0,0,0)$, $(1,1,0)$, and $(1,1,1)$. By running $100$ simulations of this network for $1000$ steps each, the following counts of revisits for the individual PBN states were obtained: $[47852$, $34724$, $17173$, $74$, $65$, $46$, $34$, $32]$. Thus, the percentages of time spent in each state are as follows: $[0.479$, $0.347$, $0.172$, $0.074$, $0.065$, $0.046$, $0.034$, $0.032]$, with the four most visited states being $(0, 0, 0)$, $(1, 1, 1)$, $(1, 1, 0)$, and $(0, 0, 1)$, first three of which with over 15\% revisiting frequency.
\end{example}

\section{Reinforcement learning}
\label{app:rl}

In this section, we provide a~concise review of reinforcement learning topics relevant in the context of our study. For a~comprehensive treatment of the subject, we refer to~\cite{SB18}. 
Reinforcement learning (RL) can be studied within a~framework based on Markov decision processes. Herein, we focus on discrete-time and finite processes defined as follows.

\begin{definition}[Markov decision process]
A~discrete time, finite Markov decision process (MDP) is a~quadruple $(\mathcal{S}, \mathcal{A}, P, R)$, where:
    \begin{itemize}
        \item $\mathcal{S}$ is a~finite set of states, referred to as the state space;
        \item $\mathcal{A}$ is the action space, i.e., a~finite set of actions that can be taken in each state;
        \item $P:\mathcal{S} \times \mathcal{A} \times \mathcal{S} \rightarrow [0,1]$ is the function of state-transition probabilities that determines the conditional probability $P_a(s, s') = \textrm{Pr}(\{S_{t+1} = s' \mid S_t = s, A_t = a\})$ of transitioning from state $s$ to $s'$ after taking action $a$ for each $s,s' \in \mathcal{S}$ and $a \in \mathcal{A}$;
        \item $R:\mathcal{S} \times \mathcal{A} \times \mathcal{S} \rightarrow \mathbb{R}$ is the reward function which determines the immediate reward $R_a(s, s')$ received after transitioning from state $s$ to $s'$ after taking action $a$ for each $s,s' \in \mathcal{S}$ and $a \in \mathcal{A}$.
    \end{itemize}
\end{definition}

The index $t \in \mathbb{Z}_{+}$ denotes a~time step from the sequence of discrete time steps $0,1,2,\ldots$. The discrete random variables $S_t$, $A_t$, and $R_t$ are the state, action, and immediate reward at time $t$, respectively. 

The RL problem is meant to be a~direct framing of the task of learning from interaction to achieve a~goal. The decision-maker and learner is referred to as the \emph{agent}. At each time step $t$, the agent interacts with the \emph{environment} which comprises everything outside the agent by implementing a~mapping $\pi:\mathcal{S} \times \mathcal{A} \rightarrow [0,1]$ from states to probabilities of selecting an~action $A_t$ in state $S_t$. This mapping is referred to as the agent's \emph{policy}, where $\pi(a \mid s)$ is the probability that $A_t=a$ given that $S_t =s$. 

Within the MDP-based framework, the RL problem is formalised as the optimal control of MDPs. RL methods specify how the agent changes its policy as a~result of its experiences with the goal of maximising the expected \emph{discounted return}:
\[
G_t \triangleq \sum_{k=0}^{\infty} \gamma^k R_{t+k+1},
\]
where $0 \leq \gamma \leq 1$ is the \emph{discount factor}.

Nearly all RL algorithms involve estimating \emph{value functions}. For MDPs, two such functions are commonly considered. First, the \emph{state-value function} $v_\pi(s)$ defined as
\[
v_\pi(s) \triangleq \mathbb{E}_\pi[G_t \mid S_t = s] = \mathbb{E}_\pi\left[ \sum_{k=0}^{\infty} \gamma^k R_{t+k+1} \bigm\vert S_t = s \right ]\!,
\]
where $\mathbb{E}_\pi[\cdot]$ is the expected value of a~random variable given that the agent follows policy $\pi$. Second, the \emph{action-value function} $q_\pi(s,a)$ is defined as the expected return starting from $s$, taking the action $a$, and thereafter following policy $\pi$:
\begin{align*}
q_\pi(s,a) &\triangleq \mathbb{E}_\pi[G_t \mid S_t = s, A_t = a]\\ &= \mathbb{E}_\pi\left[ \sum_{k=0}^{\infty} \gamma^k R_{t+k+1} \biggm\vert S_t = s, A_t = a\right]\!.
\end{align*}

The value functions satisfy particular recursive relationships expressed in the form of \emph{Bellman equations}, i.e., one for $v_\pi$:
\[
v_\pi(s) = \sum_{a \in \mathcal{A}} \pi(s,a) \sum_{s' \in \mathcal{S}} P_a(s,s')\Bigl [R_a(s,s') + \gamma v_\pi(s') \Bigr ]
\]
and one for $q_\pi$:
\[
q_\pi(s,a) = \sum_{s' \in \mathcal{S}} P_a(s,s') \Bigl [ R_a(s,s') + \gamma \sum_{a' \in \mathcal{A}} \pi (s',a') q_\pi(s',a') \Bigr].
\]

Value functions define a~partial order over policies, i.e., $\pi \geq \pi'$ if and only if $v_\pi(s) \geq v_\pi'(s)$ for all $s \in \mathcal{S}$. There always exists at least one policy which is better than or equal to all other policies~\cite{SB18}. Such policy is referred to as an~\emph{optimal policy} and all optimal policies are denoted by $\pi_*$. Optimal policies share the same \emph{optimal state-value function}, denoted $v_*$, and defined as 
\[
v_*(s) \triangleq \max_\pi v_\pi(s),
\]
for all $s \in \mathcal{S}$. They also share the same \emph{optimal action-value function}, denoted $q_*$, which is defined as
\[
q_*(s,a) \triangleq \max_\pi q_\pi(s,a),
\]
for all $s \in \mathcal{S}$. It holds that
\[
q_*(s,a) = \mathbb{E}_{\pi_*} \bigl [R_{t+1} + \gamma v_*(S_{t+1}) \bigm \vert S_t = a, A_t = a \bigr ].
\]

The optimal value functions satisfy the \emph{Bellman optimality equations}, i.e.,
\begin{align*}
v_*(s) &= \max_{a \in \mathcal{A}} \mathbb{E} \bigl [ R_{t+1} + \gamma v_*(S_{t+1}) \bigm \vert S_t= s, A_t = a\bigr]\\
&= \max_{a \in \mathcal{A}} \sum_{s' \in \mathcal{S}} P_a(s,s') \bigl [R_a(s,s') + \gamma v_*(s') \bigr ]
\end{align*}
and
\begin{align*}
q_*(s,a) &= \mathbb{E} \bigl [ R_{t+1} + \gamma\max_{a' \in \mathcal{A}} q_*(S_{t+1},a') \bigm \vert S_t= s, A_t = a\bigr]\\
&= \sum_{s' \in \mathcal{S}} P_a(s,s') \bigl [R_a(s,s') + \gamma \max_{a' \in \mathcal{A}} q_*(s',a') \bigr ],
\end{align*}
where the expectations are taken without reference to any specific policy since the value functions are optimal.

For finite MDPs, the Bellman optimality equations have unique solutions independent of the policy. If the dynamics of the MDP in known, i.e., the $P$ and $R$ functions, one can solve the Bellman optimality equations. Once the optimal state-value function $v_*$ is known, any policy that is \emph{greedy} with respect to $v_*$ is an~optimal policy. 

Optimal policies can be computed with \emph{dynamic programming} (DP) algorithms, such as \emph{policy iteration} or \emph{value iteration}, given a~perfect model of the environment in the form of an~MDP. DP algorithms are obtained by turning Bellman equations into update rules that improve the approximations of the desired value functions. However, the utility of DP algorithms is limited due to the perfect model assumption and its computational expense. Therefore, for large state space environments or cases there the model of the environment is unavailable, Monte Carlo methods or 
temporal-difference learning (TD) methods are used, see~\cite{SB18}.



\subsection{Q-learning}
\label{sec:q-learning}

Q-learning is a~TD control algorithm and is one of the most popular model-free methods for solving RL problems. It is used to learn the optimal action-selection policy. The \emph{one-step Q-learning}, which is the simplest form of Q-learning, is defined by the update rule 
\[
Q(S_t, A_t) \leftarrow Q(S_t, A_t) + \alpha \bigl[ R_{t+1} + \gamma \max_{a \in \mathcal{A}} Q(S_{t+1}, a) - Q(S_t, A_t) \bigr]\!,
\]
where $0 < \alpha \leq 1$ is the \emph{learning rate}, which determines how fast past experiences are forgotten by the agent, and the learned action-value function $Q$ directly approximates $q_*$. The algorithm can be outlined as follows. First, initialise $Q(s,a)$ for all $(s,a) \in \mathcal{S} \times \mathcal{A}$ and start the environment at state $S_0$. For time steps $t = 0, 1, 2, \ldots$, choose action $A_t$ according to the \emph{$\epsilon$-greedy policy}, i.e.,
\[
A_t \leftarrow 
\begin{cases}
\textrm{argmax}_{a \in \mathcal{A}} Q(S_t,a) & \textrm{with probability } 1 - \epsilon\\
\textrm{uniformy random action in } \mathcal{A} & \textrm{with probability } \epsilon,\\
\end{cases}
\]
and take action $A_t$ in the environment. In result, the state is changed from $S_t$ to $S_{t+1} \sim P_a(s,\cdot)$. Observe $S_{t+1}$ and $R_{t+1}$. Finally, revise the action-value function $Q$ at state-action $(S_t, A_t)$ in accordance with the above update rule.

The Q-learning algorithm iteratively updates the Q-values based on observed transitions, gradually refining its estimates of the optimal action-value function. All that is required to assure convergence to $q_*$ is that all $(s,a) \in \mathcal{S} \times \mathcal{A}$ pairs continue to be updated. Notice that the convergence is guaranteed independently of the policy being followed. This property renders Q-learning an~\emph{off-policy} TD control algorithm and enables the use of the $\epsilon$-greedy policy, which is a~simple technique for handling the exploration-exploitation tradeoff, to ensure the convergence. Once learned, the Q-values are used to provide the agent with the optimal policy for selecting actions in each state.

\section{Deep reinforcement learning}
\label{sec:Q-function_approx}

The Q-learning algorithm has proved successful for controlling small-scale environments. However, in the case of large state-action spaces, Q-learning cannot learn all action values in all states separately and can fail to converge to an~optimal solution. To address this issue, the idea was conceived to learn a~parameterised value function $Q(a,s; \theta)$ by updating the parameters $\theta$ in accordance with the standard Q-learning-derived rule:
\begin{align*}
\theta_{t+1} \leftarrow 
\theta_{t} + \alpha \Big[R_{t+1} + \gamma \max_a Q(S_{t+1}, a; \theta_t)\\
-  Q(S_t, A_t; \theta_t) \Big] \nabla_{\theta_t} Q(S_t, A_t; \theta_t),
\end{align*}
resembling stochastic gradient descent. It was shown that as the agent explores the environment, the parameterised value function converges to the true $Q$-function, see, e.g.,~\cite{SB18}. 

The above idea is realised in the deep RL (DRL) framework~\cite{MKSR+15}, which combines classical RL with artificial neural network (ANN) function approximation. In DRL, a~function approximator is trained to estimate the Q-values. The approximator is a~deep Q-network agent (DQN), i.e., a~multi-layer ANN which for a~given observed state of the environment $s$ outputs a~vector of action values $Q(s,\cdot; \theta)$, where $\theta$ are the parameters of DQN. The aim of DQN is to approximate $q_*$ by iteratively minimising a~sequence of loss functions $L_t(\theta_t)$:
\begin{align*}
L_t(\theta_t) = \left( Y_t^{\textrm{DQN}} - Q(S_t, A_t; \theta_t) \right)^2,
\end{align*}
where $Y_t^{\textrm{DQN}}$ is the DQN target defined as
\begin{align*}
Y_t^{\textrm{DQN}} \triangleq R_{t+1} + \gamma \max_a Q(S_{t+1}, a; \theta_t^-),
\end{align*}
while $\theta_t^-$ and $\theta_t$ are the parameters of the target and online networks, respectively. The target network is the same as the online network, but its parameter values are copied from the online network every $k$ steps and kept fixed at other steps.

\subsection{Double Deep Q-Network}
\label{sec:ddqn}

In the study of~\cite{MCSW22}, the Double Deep Q-Network (DDQN)~\cite{HGS16} was used to control PBNs. Both standard Q-learning and DQN uses the same values both to select and to evaluate an~action, which can lead to selection based on overestimated values~\cite{HGS16}. DDQN combines Double Q-learning~\cite{Hasselt10} with Deep Q-learning~\cite{MKSR+15} to address this issue and uses the following DDQN target 
\begin{align*}
Y_t^{\textrm{DDQN}} \triangleq R_{t+1} + \gamma Q(S_{t+1}, \textrm{argmax}_a Q(S_{t+1}, a; \theta_t);\theta_t').
\end{align*}
The second set of parameters, $\theta_t'$, enables fairly evaluation of the greedy policy. The parameters of the DDQN are updated by symmetrically switching the roles of $\theta_t$ and $\theta_t'$.

\subsection{Branching Dueling Q-Network}
\label{app:bdq}

\begin{figure*}[!hp]
  \centering
  \includegraphics[width=.8\textwidth]{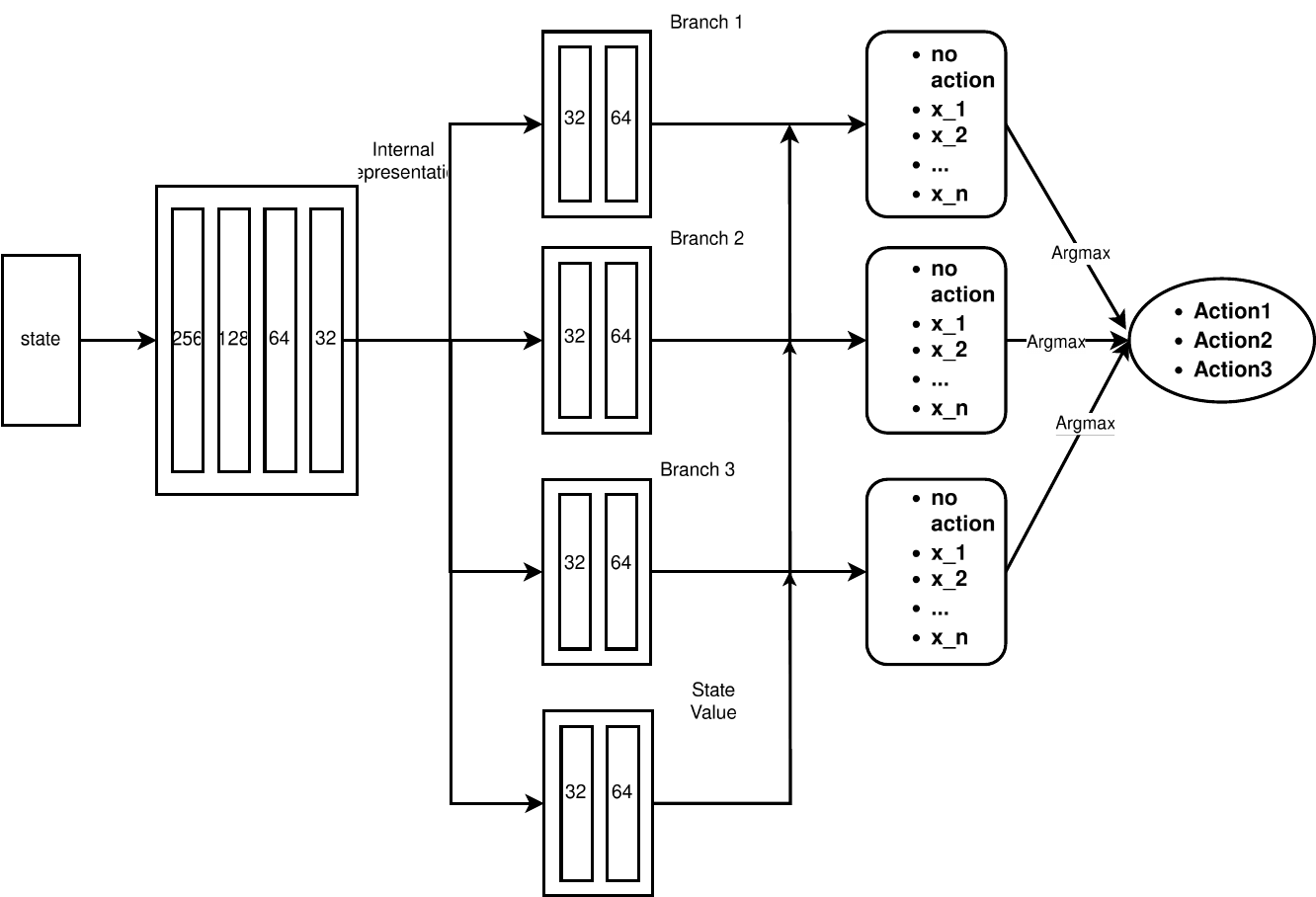}
  \caption{Schematic representation of the BDQ architecture of the DRL agent in pbn-STAC.}
  \label{fig:bdq}
\end{figure*}

In our study, to allow for simultaneous perturbation of a~combination of genes within a~DRL action, we replace the original DDQN architecture of~\cite{MCSW22} with the Branching Dueling Q-Network (BDQ) architecture~\cite{bdq}. BDQs are designed to address complex and high-dimensional action spaces; they enhance the scalability and sample efficiency of RL algorithms in complex scenarios. BDQs can be thought of as an~adaptation of the dueling network architecture~\cite{WSHH+16} into the action branching architecture. A~BDQ uses two separate ANNs, i.e., the \emph{target network} for evaluation and the \emph{controller network} for selection of actions. Furthermore, instead of using a~single output layer for all actions, BDQ introduces multiple branches, each responsible for a~specific subset of actions. 
This decomposition allows for parallel processing and shared representations, making it more efficient for the agent to learn and execute tasks with a~diverse action space. BDQs avoid overestimating Q-values, can more rapidly identify action redundancies, and generalise more efficiently by learning a~general Q-value that is shared across many similar actions. The BDQ architecture used in our study is depicted in Fig.~\ref{fig:bdq}.

\section{List of genes included in BN and PBN models of melanoma}

The genes included in individual BN and PBN models of melanoma are shown in Tab.~\ref{tab:gene_list}. 

\begin{table*}[!hp]
\centering
\caption{Genes included in individual BN and PBN models of melanoma. The presence of a~gene in a~particular model is indicated with $+$ and its absence with $-$.}
\begin{tabular}{|p{11cm}|c|c|c|}
\hline 
gene & BN/PBN-7 & BN/PBN-10 & BN/PBN-30\\
\hline\hline
reticulon 1 & +& +&+ \\
\hline
hydroxyacyl-Coenzyme A dehydrogenase/3-ketoacyl-Coenzyme A thiolase/enoyl-Coenzyme A hydratase (trifunctional protein), beta subunit &+ &+ & +\\
    \hline
ESTs &+ &+ &+ \\
\hline
pirin & +& +&+ \\
\hline
melan-A &+ & +&+ \\
\hline
wingless-type MMTV integration site family, member 5A & +& +&+ \\
\hline
S100 calcium-binding protein, beta (neural) & +& +& +\\
\hline
hexabrachion (tenascin C, cytotactin) & -& +&+ \\
\hline
CD63 antigen (melanoma 1 antigen) & -&+ & +\\
\hline
endothelin receptor type B &- &+ &+ \\
\hline
annexin A2 &- &+ & +\\
\hline
phosphofructokinase, liver &- & -& +\\
\hline
phosphoglycerate mutase 1 (brain) &- &- & +\\
\hline
synuclein, alpha (non A4 component of amyloid precursor) & -& -&+ \\
\hline
phospholipase C, gamma 1 (formerly subtype 148) &- &- & +\\
\hline
growth associated protein 43 &- &- &+ \\
\hline
AXL receptor tyrosine kinase & -&- & +\\
\hline
syndecan 4 (amphiglycan, ryudocan) &- &- & +\\
\hline
EphA2 & -&- & +\\
\hline
CD20 antigen &- &- & +\\
\hline
matrix metalloproteinase 3 (stromelysin 1, progelatinase) &- & -& +\\
\hline
integrin, beta 1 (fibronectin receptor, beta polypeptide, antigen CD29 includes MDF2, MSK12) &- &- & +\\
\hline
tropomyosin 1 (alpha) & -& -& +\\
\hline
retinoid X receptor, alpha & -& -& +\\
\hline
RAB2, member RAS oncogene family &- &- & +\\
\hline
synuclein, alpha (non A4 component of amyloid precursor) & -&- &+ \\
\hline
\end{tabular}
\label{tab:gene_list}
\end{table*}

\renewcommand\labelitemi{--}


\onecolumn

\section{Results}
\label{app:results}

\begin{figure*}[!hp]
\centering
   \subfloat[BN-7]{
     \includegraphics[width=.49\linewidth]{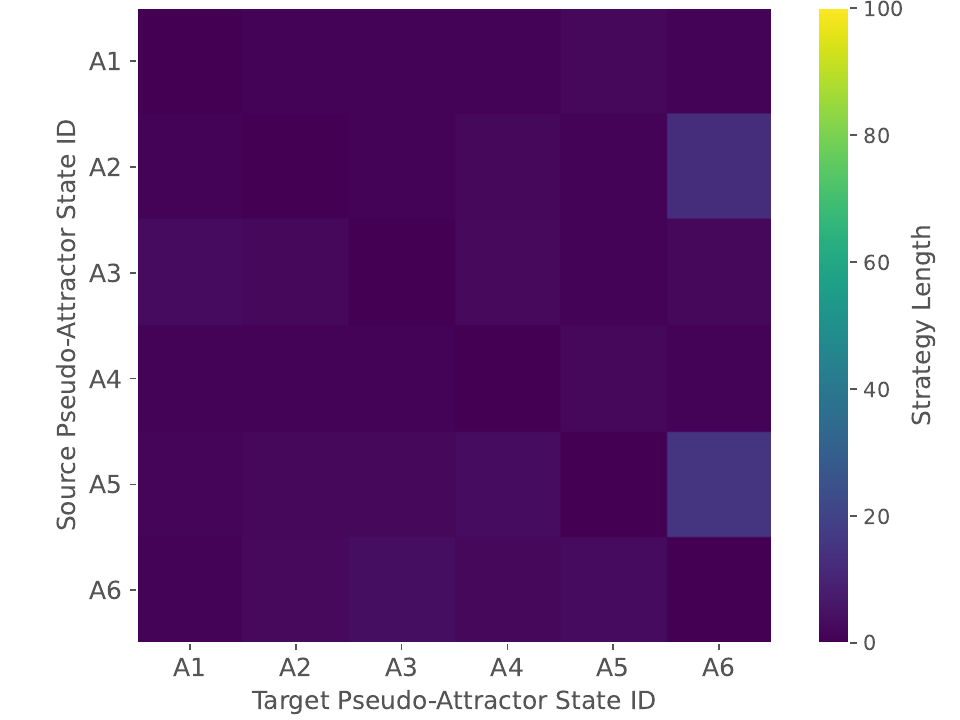}\label{fig:bn7:heat:appendix}}
    \subfloat[BN-10]{
     \includegraphics[width=.49\linewidth]{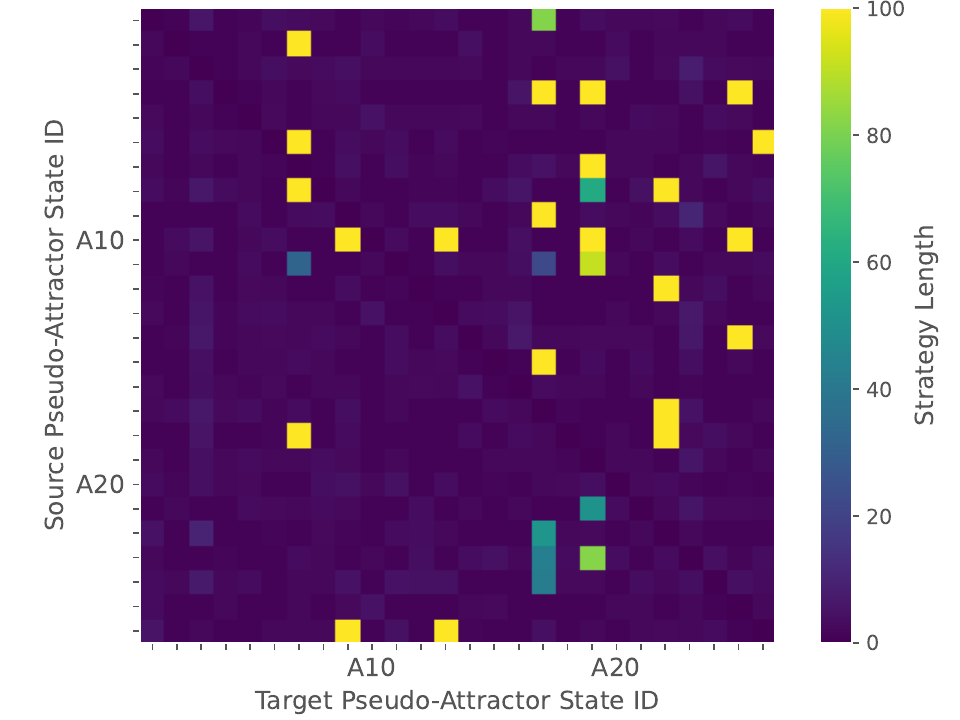}\label{fig:bn10:heat:appendix}}
     
     \subfloat[BN-30]{
     \includegraphics[width=.49\linewidth]{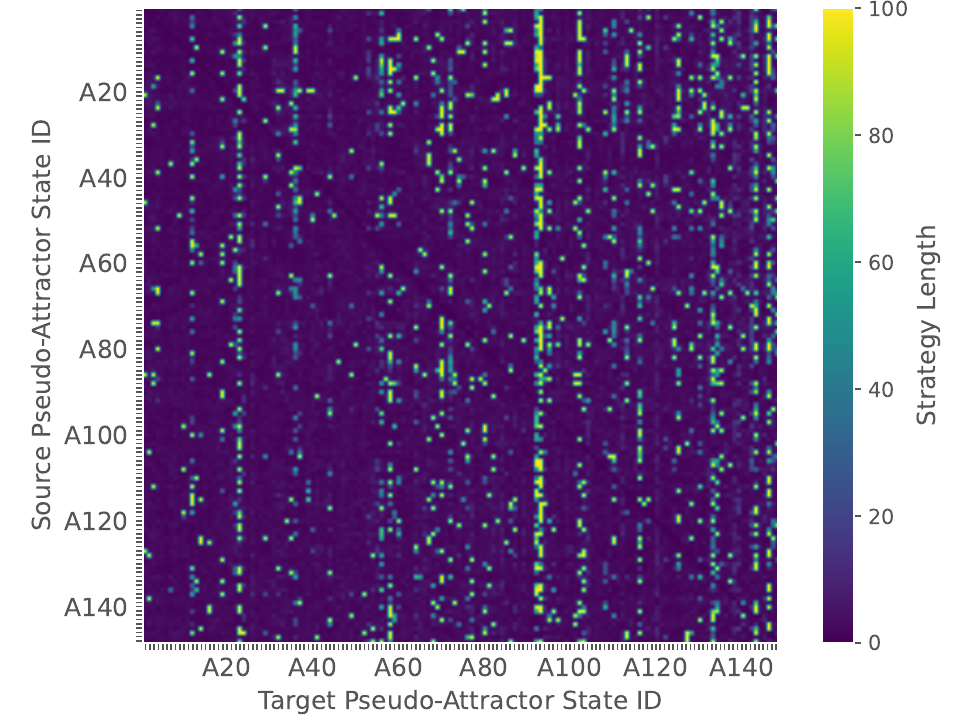}\label{fig:bn30:heat:appendix}}     
     \subfloat[IRBB-33]{
     \includegraphics[width=.49\linewidth]{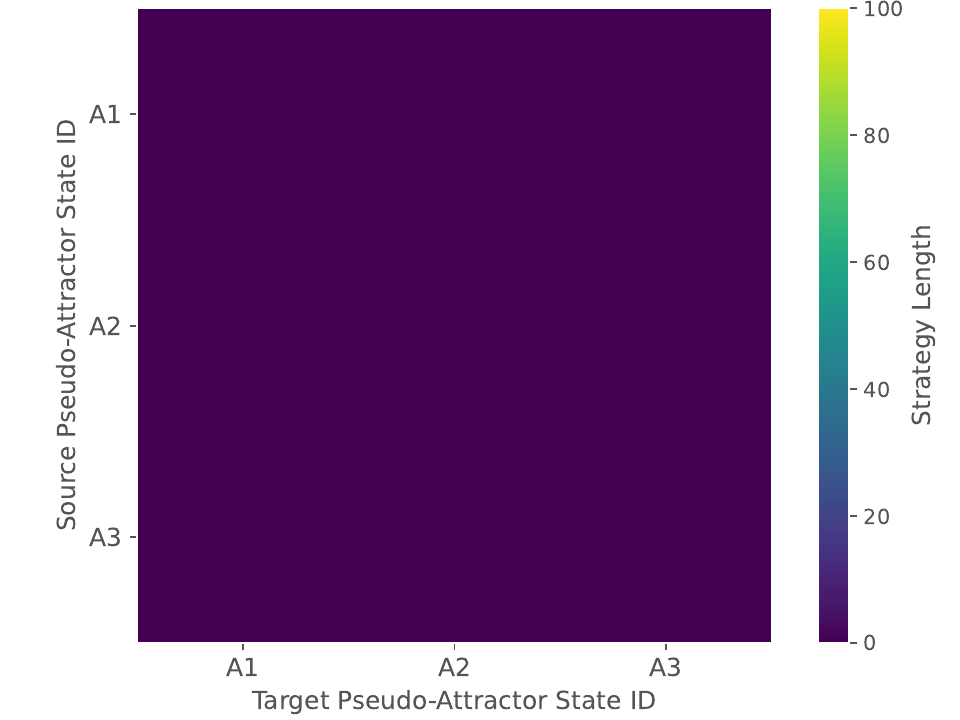}\label{fig:bn33:heat:appendix}}
\caption{Heatmaps of strategy lengths for individual source-target (pseudo-)attractor states of BN models averaged over 10 runs.}
\label{fig:model_tester_out_heat_bn:appendix}
\end{figure*}

\begin{figure*}[!hp]
\centering
  \subfloat[PBN-7]{
    \includegraphics[width=.49\linewidth]{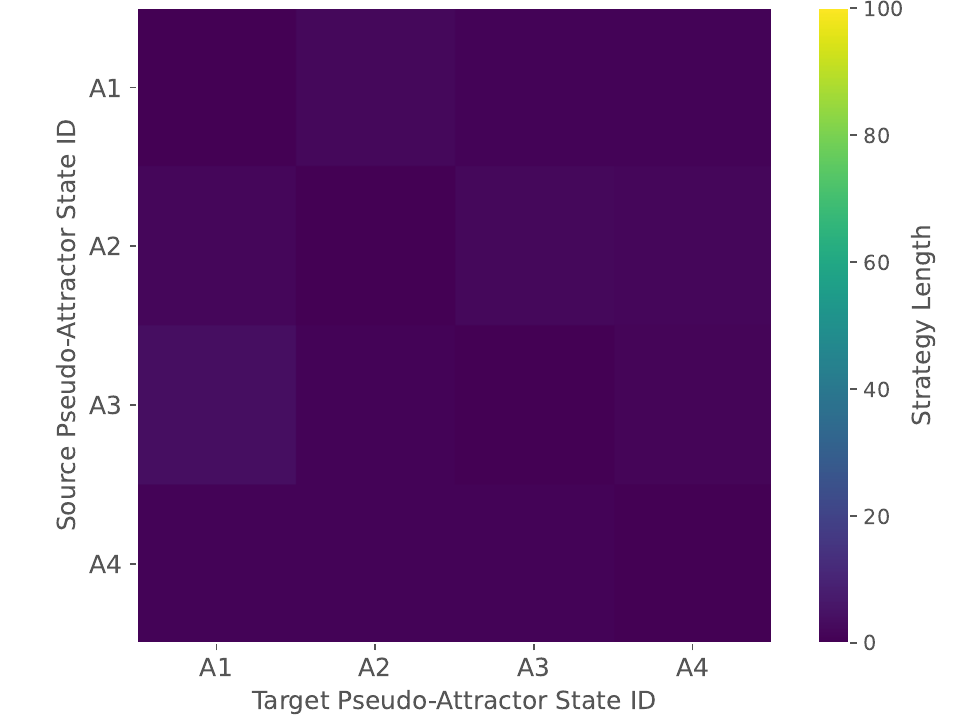}\label{fig:pbn7:heat:appendix}}
  \subfloat[PBN-10]{
    \includegraphics[width=.49\linewidth]{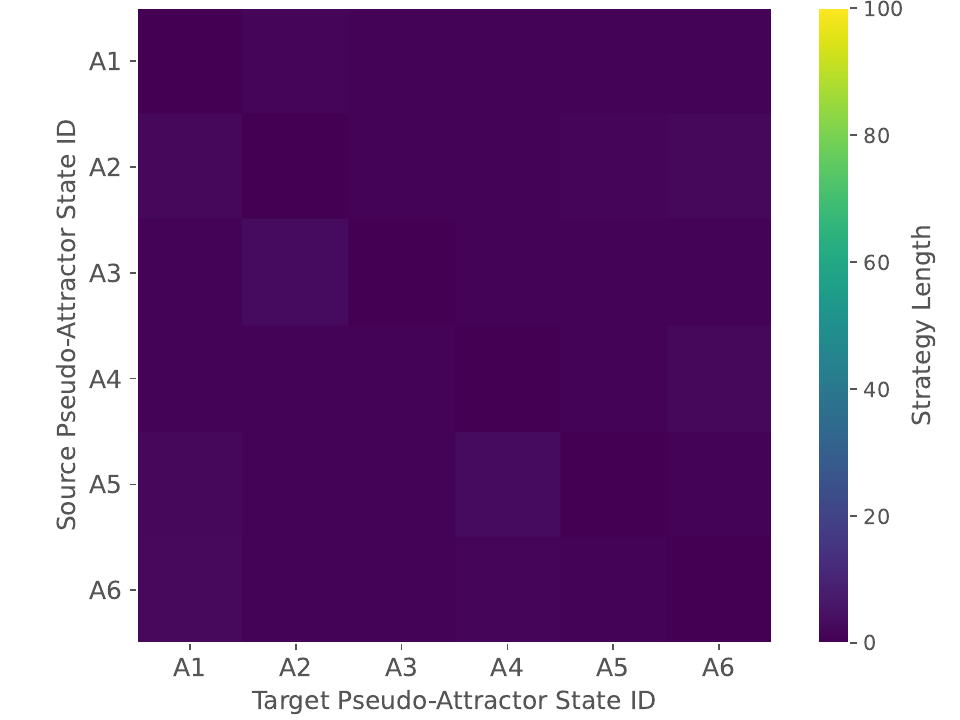}\label{fig:pbn10:heat:appendix}}

  \subfloat[PBN-30]{
  \includegraphics[width=.49\linewidth]{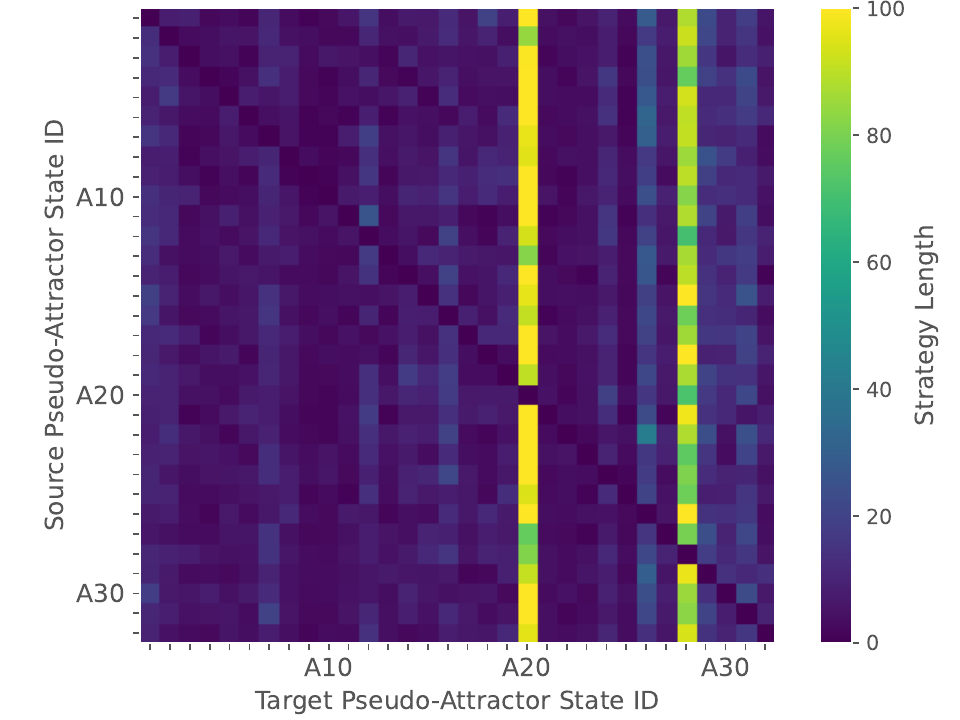}\label{fig:pbn30:heat:appendix}}   
  \caption{Heatmaps of strategy lengths for individual source-target (pseudo-)attractor states of PBN models averaged over 10 runs.}
  \label{fig:model_tester_out_heat_pbn:appendix}
\end{figure*}

\begin{figure}[!hp]
 \centering
 \subfloat[BN-7]{
    \includegraphics[width=.49\linewidth]{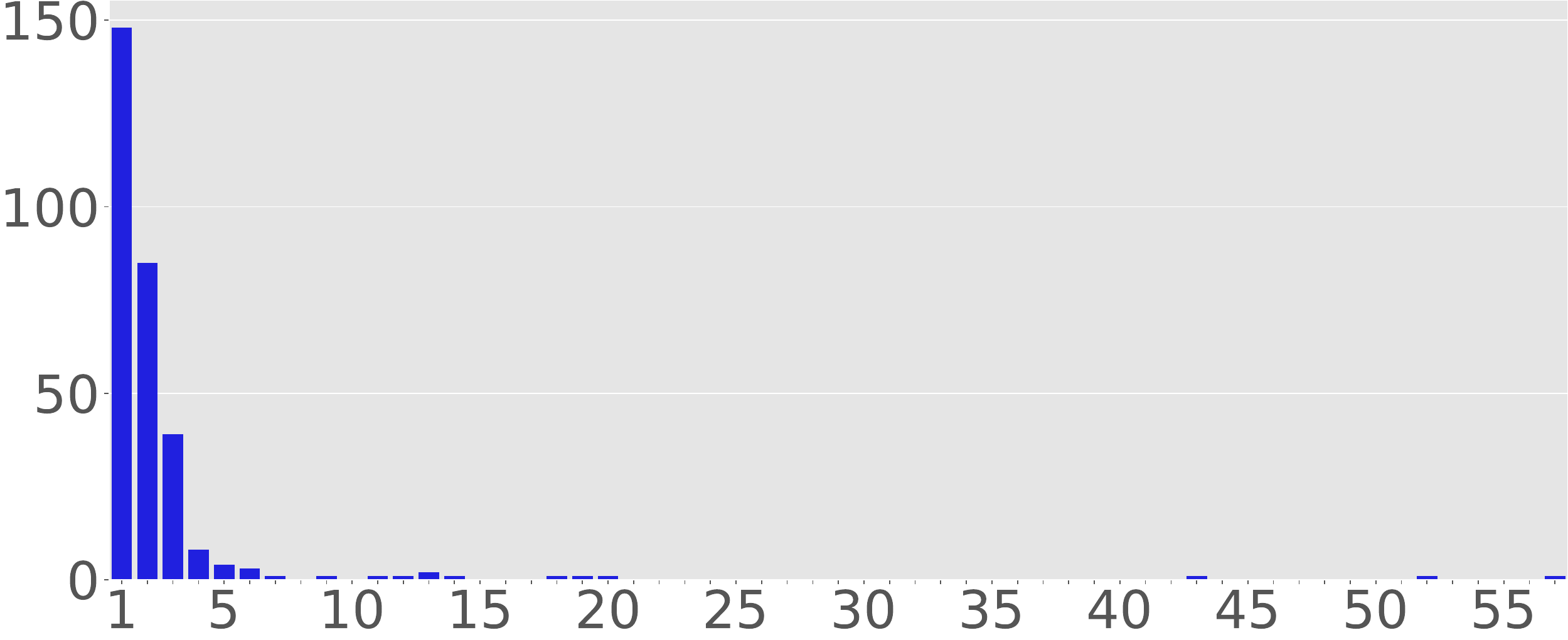}}
    \label{fig:bn7:appendix}
 \subfloat[BN-10]{
    \includegraphics[width=.49\linewidth]{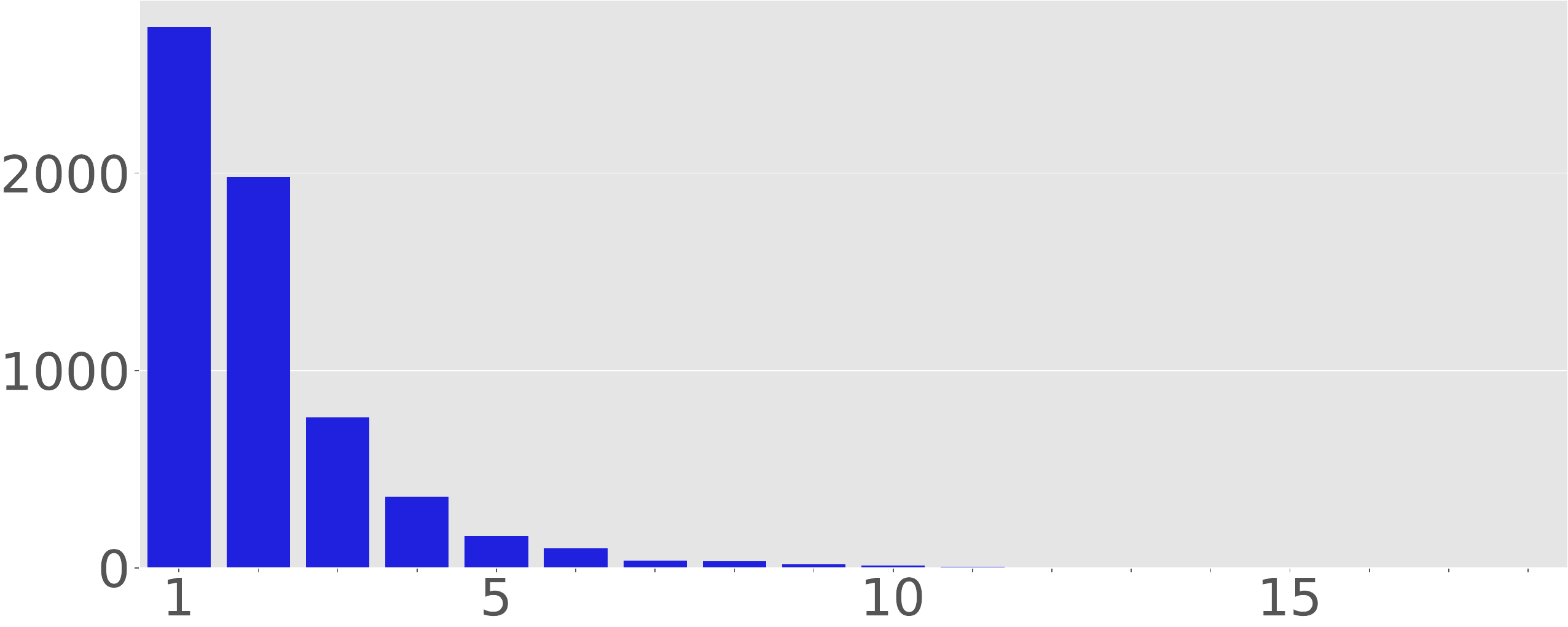}}

  \subfloat[BN-30]{
    \includegraphics[width=.49\linewidth]{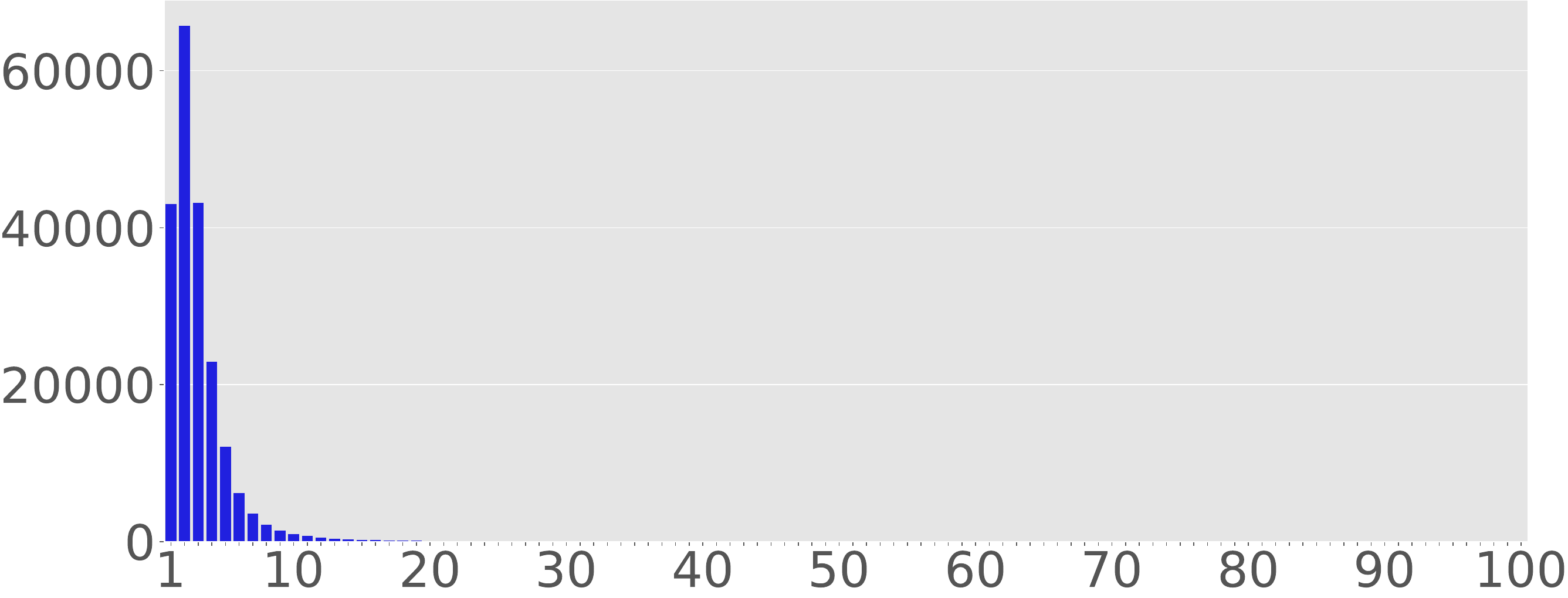}}
 \subfloat[IRBB-33]{
    \includegraphics[width=.49\linewidth]{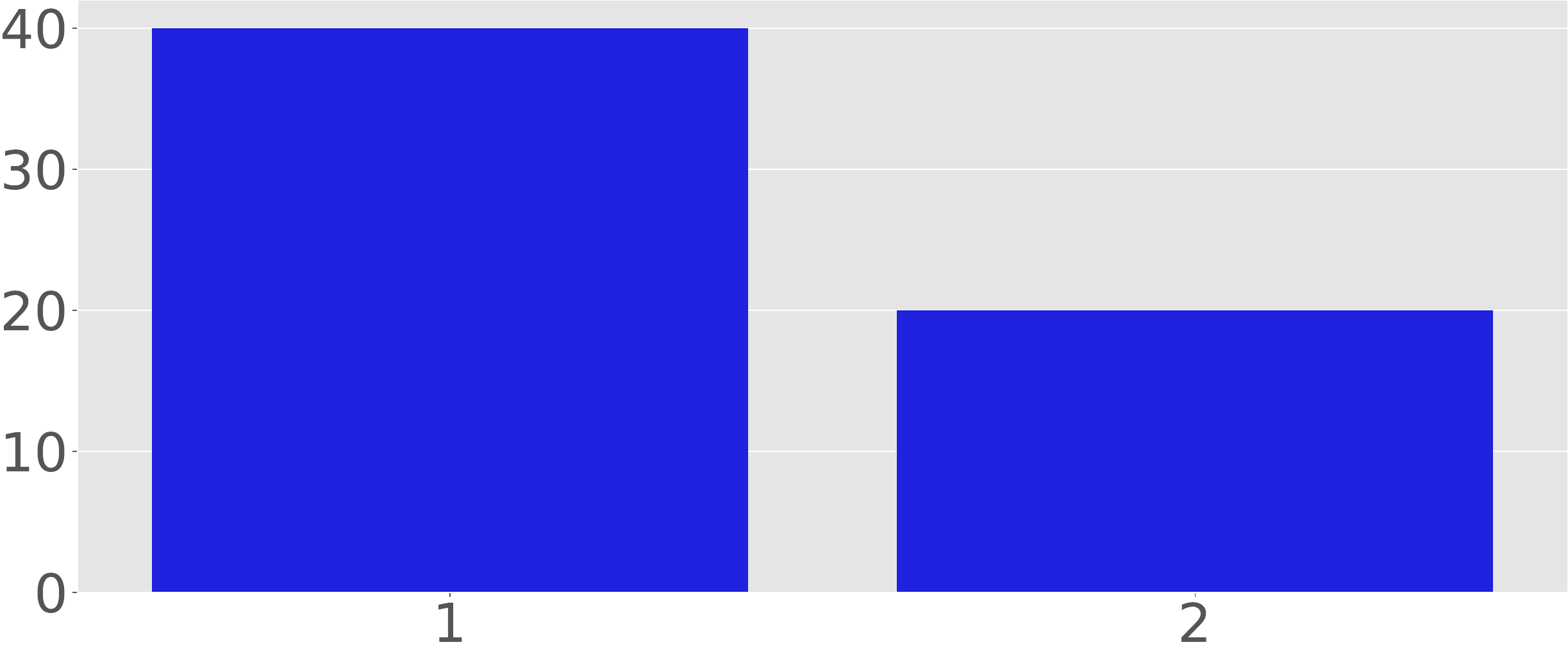}}
  
  \caption{Histograms of the number of successful control strategies vs their length for the BN models.}
  \label{fig:BN_histograms:appendix}
\end{figure}

\begin{figure}[!hp]
 \centering
 \subfloat[PBN-7]{
     \includegraphics[width=.49\linewidth]{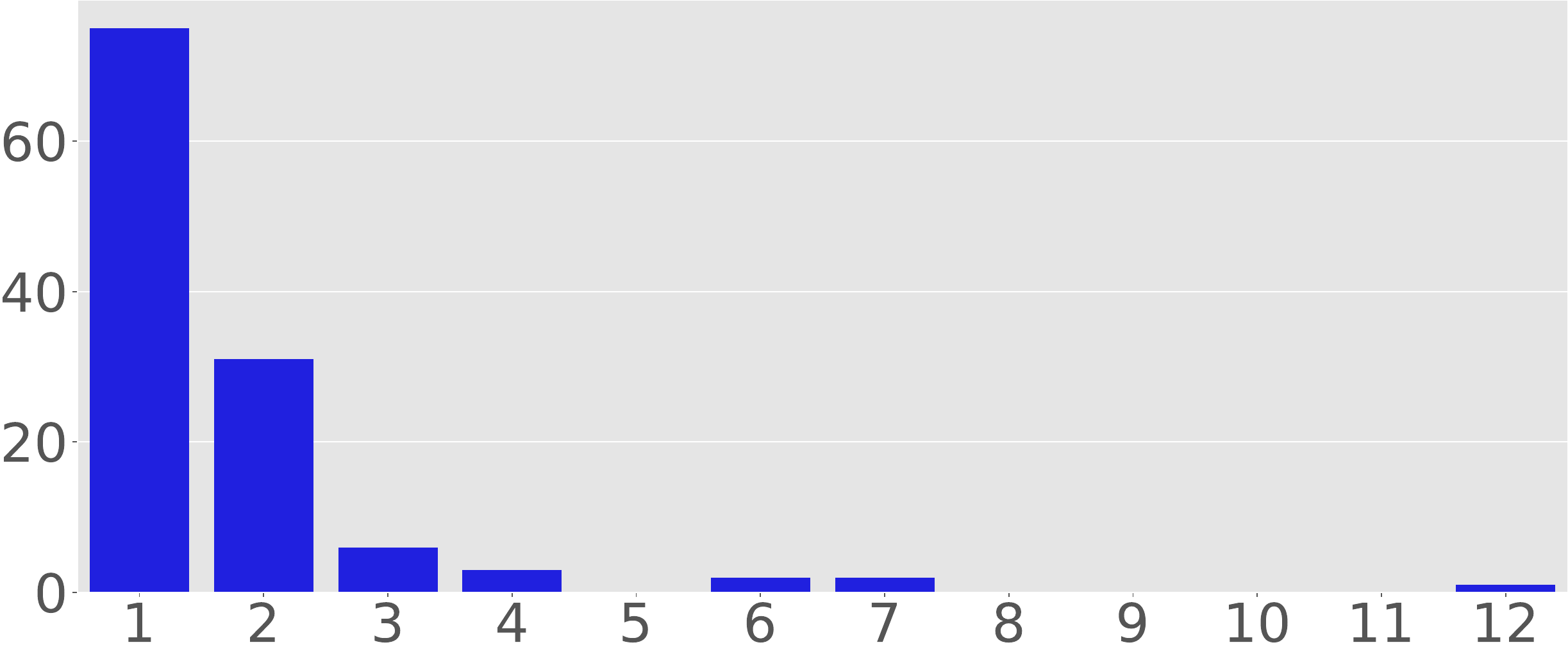},}
 \subfloat[PBN-10]{
     \includegraphics[width=.49\linewidth]{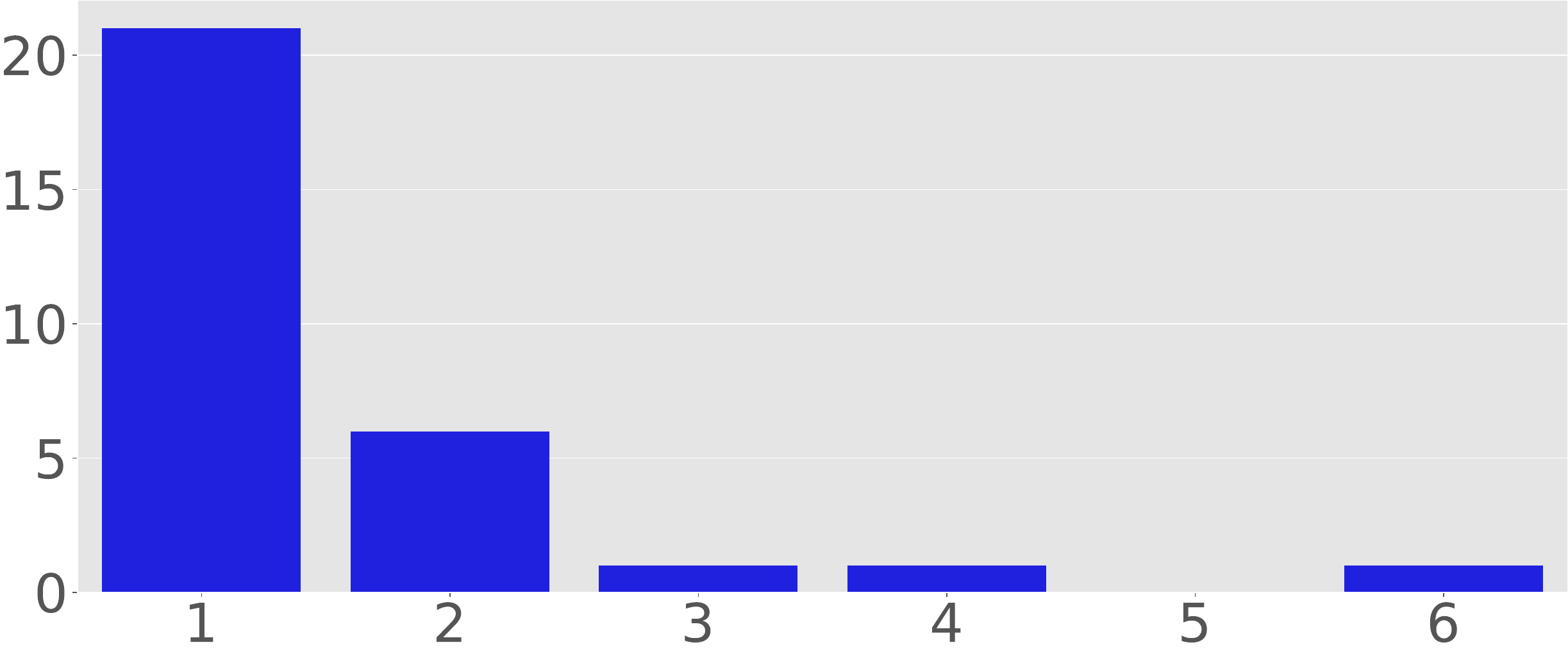}}

 \subfloat[PBN-30]{
     \includegraphics[width=.49\linewidth]{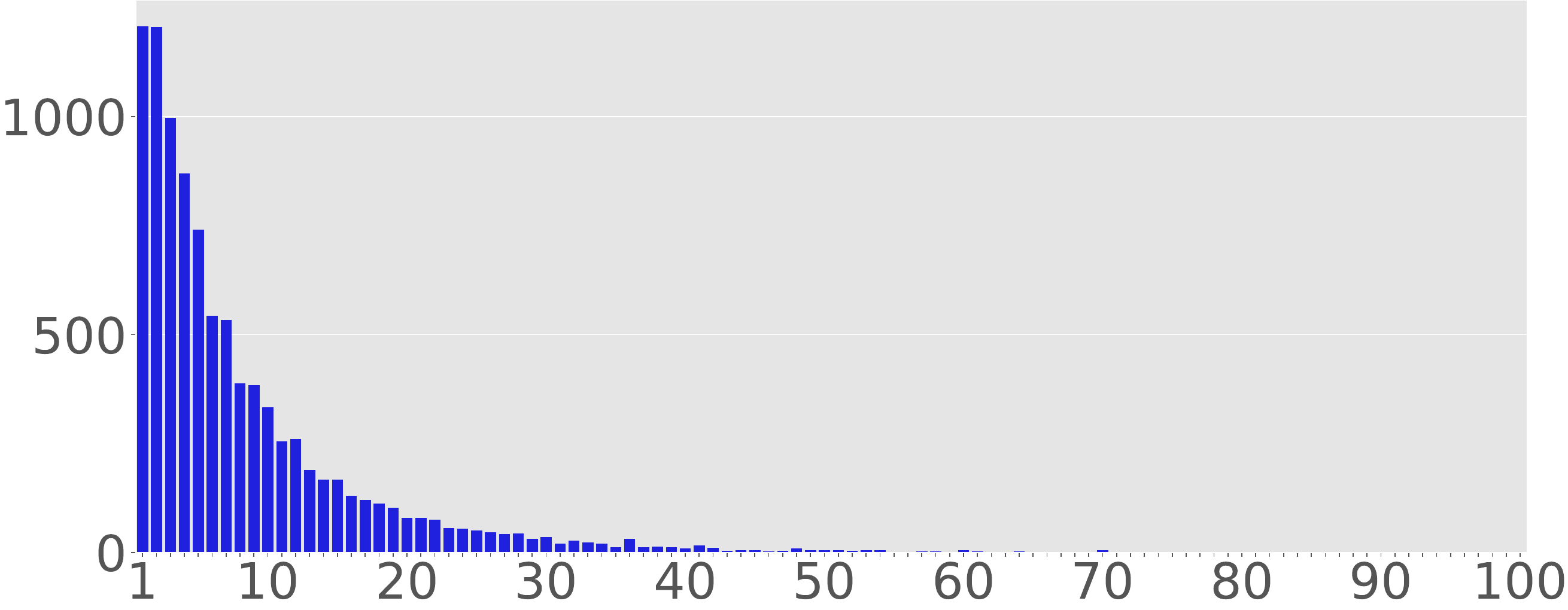}}

  \caption{Histograms of the number of successful control strategies vs their length for the PBN models.}
  \label{fig:PBN_histograms:appendix}
\end{figure}

Herein, we provide detailed information in the form of heatmaps and histograms on the control strategy lengths obtained for the considered BN models (Fig.~\ref{fig:model_tester_out_heat_bn:appendix} and Fig.~\ref{fig:BN_histograms:appendix}) and PBN models (Fig.~\ref{fig:model_tester_out_heat_pbn:appendix} and Fig.~\ref{fig:PBN_histograms:appendix}) discussed in the main text.

\end{document}